\documentclass{article}

 \usepackage[eandd, final]{neurips_2026}
\usepackage[utf8]{inputenc} % allow utf-8 input
\usepackage[T1]{fontenc}    % use 8-bit T1 fonts
\usepackage{hyperref}       % hyperlinks
\usepackage{url}            % simple URL typesetting
\usepackage{booktabs}       % professional-quality tables
\usepackage{amsfonts}       % blackboard math symbols
\usepackage{nicefrac}       % compact symbols for 1/2, etc.
\usepackage{microtype}      % microtypography
\usepackage{xcolor}         % colors

\usepackage{color, colortbl}
\usepackage{graphicx}
\usepackage{wrapfig}

\usepackage{algorithm}
\usepackage{algorithmic}
\usepackage{adjustbox}
\usepackage{multirow}
\usepackage{pifont}
\usepackage{comment}
\usepackage{subcaption}
\usepackage{array}
\usepackage{tikz} 
\usepackage{acronym}
\usepackage{xspace}
\usepackage{enumitem}
\usepackage{tabularx}
\usepackage{threeparttable}

\usepackage{fontawesome5}

\definecolor{ProjectBlue}{HTML}{2563EB}
\definecolor{GitHubPurple}{HTML}{7C3AED}
\definecolor{HFOrange}{HTML}{D97706}

\newcommand{\hficon}{%
  \raisebox{-0.15em}{%
    \includegraphics[height=1.05em]{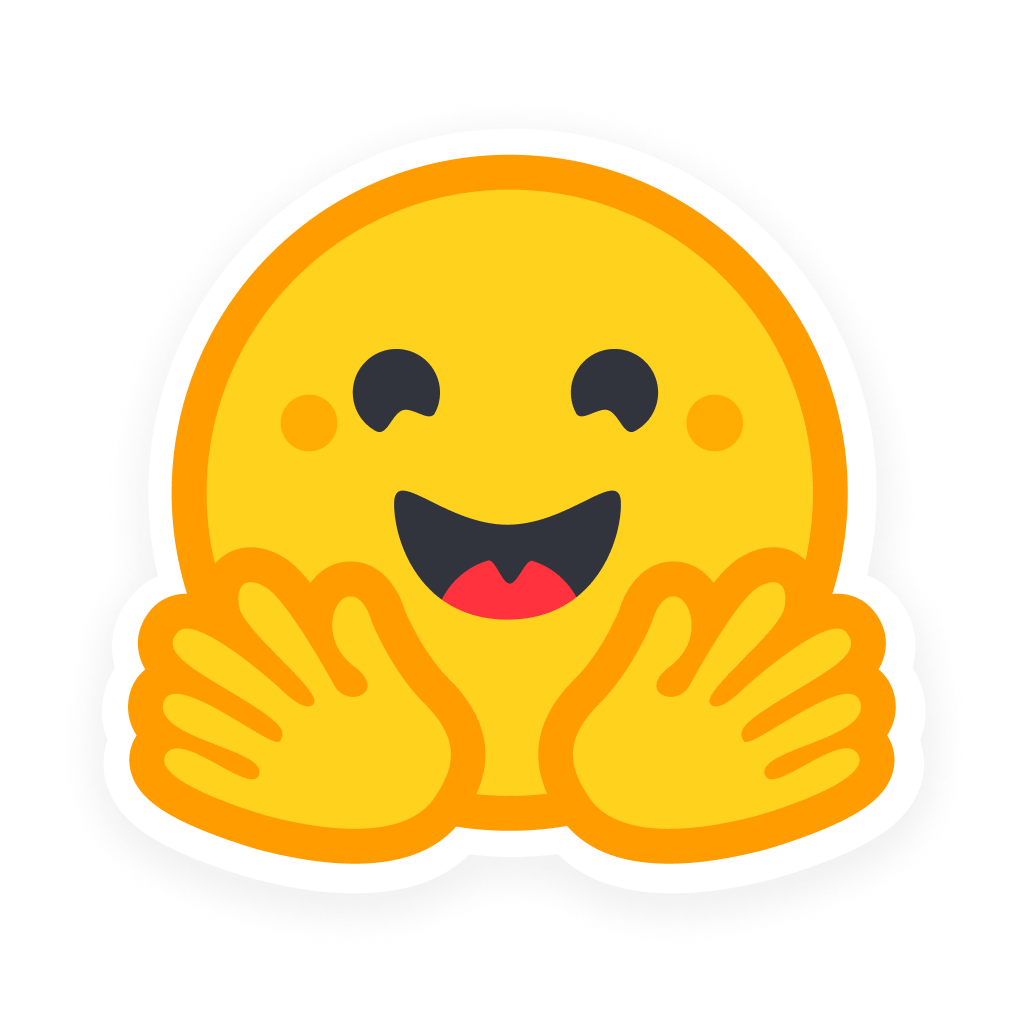}%
  }%
}

\newcommand{\projectlinks}{%
  \begingroup
  \hypersetup{hidelinks}%
  \href{https://kangverse.github.io/GameVerse/}{%
    \textcolor{ProjectBlue}{%
      \faGlobe\enspace\textbf{Project Page}%
    }%
  }%
  \qquad
  \href{https://github.com/kangverse/GameVerse}{%
    \textcolor{GitHubPurple}{%
      \faGithub\enspace\textbf{GitHub}%
    }%
  }%
  \qquad
  \href{https://huggingface.co/datasets/Kangverse/GameVerse}{%
    \hficon\enspace
    \textcolor{HFOrange}{\textbf{Datasets}}%
  }%
  \endgroup
}

\definecolor{Color1}{gray}{0.95}
\definecolor{Color2}{gray}{0.87}
\definecolor{methodcolor}{HTML}{8CE4FF}
\definecolor{mygreen}{rgb}{0.032, 0.6392, 0.2039}
\definecolor{nmgray}{RGB}{229,229,229}
\definecolor{mycolor}{HTML}{FF5F00}
\definecolor{delta_color}{HTML}{16C47F}
\definecolor{kangColor}{rgb}{1,0.33,0.64}

\definecolor{gvL}{HTML}{6366F1} % indigo
\definecolor{gvR}{HTML}{EC4899} % pink
\newcommand{\GVName}{%
  \textcolor{gvL!100!gvR}{G}%
  \textcolor{gvL!88!gvR}{a}%
  \textcolor{gvL!75!gvR}{m}%
  \textcolor{gvL!62!gvR}{e}%
  \textcolor{gvL!50!gvR}{V}%
  \textcolor{gvL!38!gvR}{e}%
  \textcolor{gvL!25!gvR}{r}%
  \textcolor{gvL!12!gvR}{s}%
  \textcolor{gvL!0!gvR}{e}%
}

\newcommand{\cmark}{\textcolor{mygreen}{\ding{51}}}%
\newcommand{\xmark}{\textcolor{gray}{\ding{55}}}%
\newcommand{\cmarkb}{\textcolor{green!60!black}{\textbf{\ding{51}}}} 

\usepackage{CJKutf8}

\makeatletter
\DeclareRobustCommand\onedot{\futurelet\@let@token\@onedot}
\def\@onedot{\ifx\@let@token.\else.\null\fi}
\def\eg{\emph{e.g}\onedot}

\makeatother

\title{\GVName: A Minute-Scale Gameplay Dataset for Long-Horizon Interactive World Modeling}

\author{%
  Kang He$^{1,2,3,\dagger}$,
  Wenshuo Peng$^{4,\dagger}$,
  Chuanhao Li$^{1}$,
  Zihui Gao$^{1}$,
  Xiaojie Xu$^{1}$,
  Zhengyuan Lin$^{1}$, \\
  \textbf{Yuzhe Ding}$^{3}$,
  \textbf{Donghong Ji}$^{3}$,
  \textbf{Kaipeng Zhang}$^{1,2,\ast,\ddagger}$,
  \textbf{Yongtao Ge}$^{1,\ast,\ddagger}$ \\
  $^{1}$Alaya Lab \quad
  $^{2}$Shanghai Innovation Institute \\
  $^{3}$School of Cyber Science and Engineering, Wuhan University \quad
  $^{4}$Tsinghua University \\
  \texttt{hekang0225@whu.edu.cn, kaipeng.zhang@shanda.com} \\[0.3em]
  \projectlinks
}

\begin{document}

\maketitle

\begingroup
\renewcommand{\thefootnote}{}
\footnotetext{%
  $^{\dagger}$Equal contribution, work done during internship at Alaya Lab. \quad 
  $^{\ast}$Corresponding authors.\quad
  $^{\ddagger}$Project leaders.
}
\endgroup

\begin{figure}[h]
\vspace{-8mm}
    \centering
    \includegraphics[width=\linewidth]{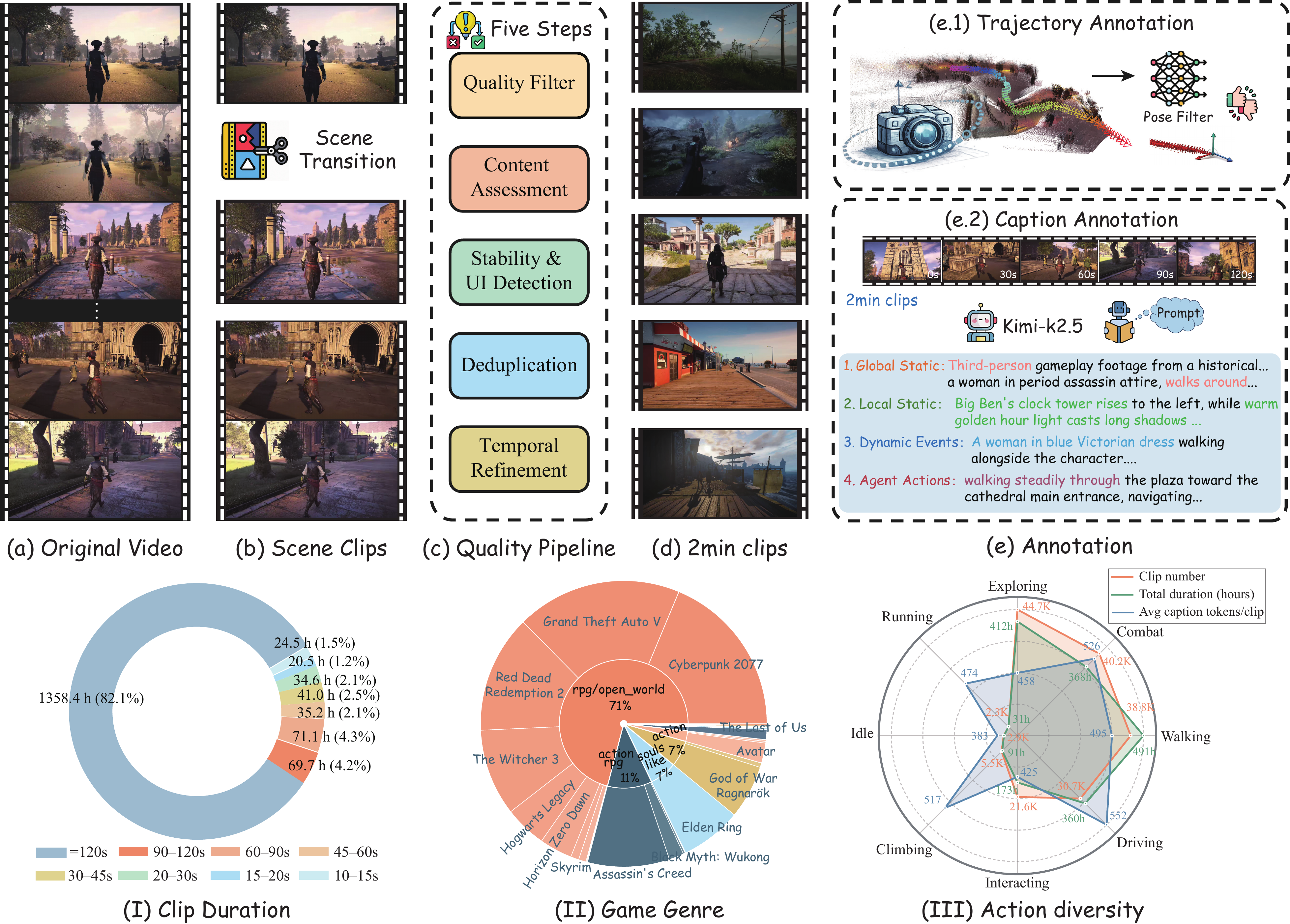}
    \caption{
    Overview of the proposed \textbf{GameVerse} dataset and construction pipeline. The top portion (a)--(e) illustrates the construction pipeline, with panel~(e.2) detailing the four-level structured caption schema (\emph{global static}, \emph{local static}, \emph{dynamic events}, \emph{agent actions}). The bottom panels present statistical analyses of GameVerse: (I)~clip-duration distribution, (II)~game-category composition, and (III)~interaction diversity across gameplay behaviors.
}
\label{fig:dataset_gameverse}
\vspace{-3mm}
\end{figure}

\begin{abstract}
Recent advances in interactive world models have made it possible to generate controllable, dynamic environments from action inputs. Yet current models struggle to maintain coherence beyond roughly one minute of rollout—a limitation rooted not in model capacity, but in the scarcity of long-horizon training data. Existing gameplay video datasets typically consist of clips under ten seconds, with captions that describe whole clips in coarse and undifferentiated text, which is insufficient for controllable long-horizon generation.
We address this with \textbf{GameVerse}, a minute-scale gameplay video dataset purpose-built for long-horizon interactive world modeling. GameVerse contains 69.7K clips spanning 1,655 hours and 349.6 million frames across diverse first- and third-person gameplay, including \textbf{1,358 hours of continuous two-minute sequences}—an order of magnitude much longer than prior datasets. Clips pass through a unified pipeline of scene-consistent segmentation, quality filtering, camera pose estimation, and event-based caption annotation. Each clip is paired with a temporally stable camera trajectory that remains coherent across minutes, together with captions that separate static scene context from dynamic foreground events and align to event boundaries rather than fixed time windows.
Comprehensive analysis shows that GameVerse surpasses existing public datasets in scale, diversity, and annotation quality. Experiments further demonstrate that training on GameVerse substantially extends the temporal horizon of interactive world models. 
\end{abstract}

\section{Introduction}
\label{sec:intro}

Interactive world models~\cite{bruce2024genie,DBLP:conf/iclr/ValevskiLAF25,xiaoworldmem} aim to learn controllable, dynamic environments that can be rolled out from action and camera inputs, a capability central to embodied decision-making, simulation, and game-style content generation. Recent gameplay-oriented systems such as GameGen-X~\cite{chegamegen}, Matrix-Game~\cite{zhang2025matrix}, Hunyuan-GameCraft~\cite{li2025hunyuan}, and HY-World~1.5~\cite{hyworld2025} demonstrate compelling short-horizon controllability, yet they share a striking ceiling: rollouts collapse into drift, identity loss, and scene incoherence after roughly one minute of generation. We argue that this ceiling is, first and foremost, a data problem rather than a model-capacity problem. The training corpora behind these systems consist almost entirely of clips shorter than ten seconds, paired with whole-clip captions that conflate static context with dynamic events. Such data fundamentally cannot supervise minute-scale temporal coherence, fine-grained controllability, or event-level semantics, no matter how large the model.

This limitation is visible across recent gameplay and exploration datasets, which exhibit a consistent pattern of single-axis optimization (Table~\ref{tb:intro_compare}). Sekai~\cite{lisekai,he2026sekai2} provides large-scale exploration video but lacks dense interactions, validated camera trajectories, and structured semantics. Matrix-Game and HY-World~1.5 expose engine-level camera poses but cap continuous footage at short windows. GameGen-X and Hunyuan-GameCraft support action-conditioned interaction but lack reliable trajectories and event-aligned captions. None of them simultaneously provides minute-scale continuous footage, validated frame-level camera trajectories, dense first- and third-person interaction coverage, and multi-granular event-aligned semantic annotations. So long as these four signals remain fragmented across separate datasets, any model trained on a single source inherits its blind spot.

We close this gap with \textbf{GameVerse}, a minute-scale gameplay dataset purpose-built for long-horizon interactive world modeling. Curated from 11{,}002 source videos at 1080p and 30--60\,FPS through a unified pipeline of scene-consistent segmentation, multi-stage quality filtering, ViPE-based~\cite{huang2025vipe} pose estimation with trajectory validation, and event-aligned multi-granular captioning, GameVerse delivers \textbf{69.7K clips}, \textbf{1{,}655 hours}, and \textbf{349.6M frames}, including \textbf{1{,}358 hours of continuous two-minute sequences}---roughly an order of magnitude longer than any prior gameplay or exploration dataset. Every clip is paired with a temporally stable frame-level camera trajectory that survives explicit jump-, jitter-, and drift-filtering; a four-level structured caption (\emph{global static}, \emph{local static}, \emph{dynamic events}, \emph{agent actions}) that disentangles scene context from foreground change; and dense behavior coverage across first- and third-person perspectives, spanning exploration, combat, walking, driving, climbing, and complex object interaction. To our knowledge, GameVerse is the first dataset to ship these four signals jointly under a single quality bar, and fine-tuning a strong video generator on it improves VBench~\cite{huang2024vbench} temporal-consistency and visual quality, and WorldScore~\cite{duan2025worldscore} content alignment and 3D consistency, over gameplay-domain baselines on a 30-second benchmark.

Our main contributions are summarized as follows:
\begin{itemize}[leftmargin=*,topsep=0.3em,itemsep=0.2em]
    \item \textbf{The largest minute-scale gameplay dataset to date.} 69.7K clips, 1{,}655 hours, and 349.6M frames, including 1{,}358 hours of continuous two-minute sequences---roughly an order of magnitude longer than existing gameplay corpora and the first to directly supervise the minute-scale rollout regime where current interactive world models fail.
    \item \textbf{A unified construction pipeline with reliable camera trajectories.} Scene-consistent segmentation that combines hard-cut detection with CLIP-based gradual-transition detection, multi-stage visual / content / temporal quality filtering, and ViPE-based pose estimation with explicit jump-, jitter-, and drift-filtering—delivering temporally stable trajectories at minute scale, where most prior datasets either omit poses or expose noisy ones.
    \item \textbf{A multi-granular event-aligned caption schema.} A four-level structured annotation (\emph{global static}, \emph{local static}, \emph{dynamic events}, \emph{agent actions}) that disentangles environment context from controllable behavior and supports hierarchical conditioning—replacing the single-line whole-clip captions that dominate prior gameplay data.
    \item \textbf{Empirical validation on long-horizon generation.} Training on GameVerse improves VBench temporal-consistency and visual-quality metrics and most WorldScore controllability metrics over gameplay-domain baselines on a held-out long-duration benchmark.
\end{itemize}

\begin{table}[t]
\caption{
Comparison with recent world modeling datasets.
\textbf{Long-duration} refers to minute-level continuous video sequences.
\textbf{Long-horizon} denotes the presence of temporally coherent state evolution and interaction dynamics over extended time.
\textbf{Pose} indicates the availability of \emph{publicly released}, reliable camera trajectories ($\sim$ marks systems that obtain camera poses internally but do not release them as supervision).
\textbf{Multi-gran.} denotes multi-level semantic decomposition of scene and behavior.
$\sim$ indicates partial support or not explicitly specified.
}
\label{tb:intro_compare}
\centering

\vspace{2mm}

{
\setlength{\tabcolsep}{4pt}
\renewcommand{\arraystretch}{1.15}

\resizebox{\textwidth}{!}{
\begin{tabular}{l|ccccccc}
\toprule
 & Sekai~\cite{lisekai} & GameGen-X~\cite{chegamegen} & WorldMem~\cite{xiaoworldmem} & Matrix-Game~\cite{zhang2025matrix,he2025matrix,skywork2026matrix3} & Hunyuan-GameCraft~\cite{li2025hunyuan} & HY-World~\cite{hyworld2025}  & \textbf{Ours} \\
\midrule
\midrule

Long-duration 
& $\sim$ & \xmark & \xmark & $\sim$ & \xmark & $\sim$ & \cmarkb \\

Long-horizon
& \xmark & $\sim$ & \xmark & \cmark & \xmark & \cmark & \cmarkb \\

Pose
& $\sim$ & \xmark & $\sim$ & $\sim$ & \xmark & $\sim$ & \cmarkb \\

Multi-gran. 
& \xmark & $\sim$ & \xmark & \xmark & \xmark & \xmark & \cmarkb \\

Interaction 
& \xmark & \cmark & \xmark & \cmark & \cmark & \cmark & \cmarkb \\

% Pose Align. 
% & $\sim$ & \xmark & \cmark & \cmark & \xmark & \cmark & \cmarkb \\

\bottomrule
\end{tabular}
}
}
\vspace{-3mm}
\end{table}

\section{Related Work}

\subsection{Interactive World Models and Gameplay Datasets}

Early work on interactive video generation established the methodological foundation but relied on closed, narrow training corpora. Genie~\cite{bruce2024genie} learns latent actions from an unreleased, curated 30K-hour platformer collection, leaving the community without a reusable substrate. GameNGen~\cite{DBLP:conf/iclr/ValevskiLAF25} and Oasis~\cite{decart2024oasis} demonstrate that diffusion models can serve as real-time game engines, yet rely on a single game (\eg DOOM) or a closed Decart corpus, limiting cross-game generalization and broader scientific study. WorldMem~\cite{xiaoworldmem} introduces an explicit memory module for long-term Minecraft simulation, showing that long-horizon coherence is methodologically tractable, but remains limited to a narrow evaluation domain without a transferable, large-scale gameplay dataset.

More recent gameplay-oriented systems advance controllability with action and camera signals~\cite{team2026alayaworld,yin2026alaya}, but fall short on the long-horizon \emph{data} axis. GameGen-X~\cite{chegamegen} is trained predominantly on gameplay clips of a few seconds, providing little supervision for minute-scale rollouts. The Matrix-Game family---the Minecraft-centered Matrix-Game~\cite{zhang2025matrix} and Matrix-Game~2.0~\cite{he2025matrix}, and the recent Matrix-Game~3.0~\cite{skywork2026matrix3} targeting long-horizon coherence via memory-augmented diffusion on an industrial-scale Unreal-Engine and AAA-game data engine---releases model weights and code but keeps \emph{its gameplay training corpora and Video-Pose-Action-Prompt quadruplet data proprietary}, leaving no shared substrate for scaling, ablation, and reproducibility. Our work complements Matrix-Game~3.0: while it addresses the long-horizon problem from the model side through memory on a closed data engine, GameVerse addresses it from the data side with an openly released, multi-game, minute-scale corpus that any model can consume. Hunyuan-GameCraft~\cite{li2025hunyuan} and HY-World~1.5~\cite{hyworld2025} report strong qualitative results across diverse environments but rely on unreleased internal gameplay recordings; WorldPlay~\cite{sun2025worldplay} pursues real-time long-horizon generation under the same proprietary-data constraint, while VMem~\cite{li2025vmem} explores surfel-indexed memory but remains bound to a method-specific corpus. Across this landscape, the lack of a public, diverse, multi-game, \emph{minute-scale} gameplay dataset remains the principal bottleneck, rather than model capacity.

\subsection{Camera-Aware Video Supervision}

Explicit camera control is a key axis of controllable video generation~\cite{xu2026worldmark}. On the method side, CameraCtrl~\cite{he2024cameractrl} and MotionCtrl~\cite{wang2024motionctrl} demonstrate that trajectory conditioning materially improves geometric consistency in generated video, but both rely on the data side to provide reliable trajectories for training. On the data side, ViPE~\cite{huang2025vipe} provides scalable monocular pose estimation from unconstrained video and is the de-facto tool for retrofitting trajectories onto large video corpora; we leverage it as a tool in our pipeline rather than as a competing dataset. Among existing datasets, Matrix-Game~\cite{zhang2025matrix,he2025matrix,skywork2026matrix3}, Hunyuan-GameCraft~\cite{li2025hunyuan}, and HY-World~1.5~\cite{hyworld2025} obtain engine-level camera poses in their closed training pipelines, which are never released as supervision; Sekai~\cite{lisekai,he2026sekai2} ships estimated trajectories alongside long-duration walking video but does not validate them against sudden jumps, high-frequency jitter, or long-term drift, leaving its trajectories noisy in practice at the minute scale; and the bulk of general-purpose video-language datasets ship no calibrated poses at all. GameVerse instead pairs every clip with a frame-level ViPE trajectory that has passed explicit jump-, jitter-, and drift-filtering, delivering minute-scale trajectories that remain physically coherent across the full clip, and we release these trajectories openly together with the video sources.

\subsection{Structured Caption Annotations for Video}

Caption supervision drives controllable generation, but existing video datasets stop short of the structure required for long-horizon and behavior-conditioned synthesis. HowTo100M~\cite{miech2019howto100m} relies on ASR transcripts that are loosely aligned and dominated by narration rather than scene description. InternVid~\cite{wanginternvid} and ActivityNet-style dense-captioning data~\cite{krishna2017dense,DBLP:conf/cvpr/HeilbronEGN15} contribute auto-generated or sparse event captions but treat each clip as a flat sequence of events, with no separation of static scene context from dynamic foreground and no isolation of agent behavior from environmental change. MiraData~\cite{ju2024miradata} pushes toward longer-duration video with longer structured captions, but its semantic schema is tuned to real-world video and does not include an agent-action vocabulary suitable for interactive generation. On the gameplay side, GameGen-X~\cite{chegamegen} provides single-line clip-level captions that conflate scene context with dynamic events, and Sekai~\cite{lisekai} similarly attaches one whole-clip description to each minute-long video; in either case the controllable channel cannot be modulated independently of the scene. Matrix-Game~\cite{zhang2025matrix,he2025matrix,skywork2026matrix3}, Hunyuan-GameCraft~\cite{li2025hunyuan}, and HY-World~1.5~\cite{hyworld2025} do not publicly release caption supervision at all. GameVerse instead annotates every clip with a four-level hierarchical schema---\emph{global static}, \emph{local static}, \emph{dynamic events}, and \emph{agent actions}---generated by a vision-language model~\cite{team2026kimi} and aligned to event boundaries rather than fixed time windows, so that environment context is disentangled from controllable agent behavior and can be used directly as a hierarchical conditioning signal.

\section{Dataset Construction Pipeline}

\begin{figure}[t]
    \centering
    \includegraphics[width=\linewidth]{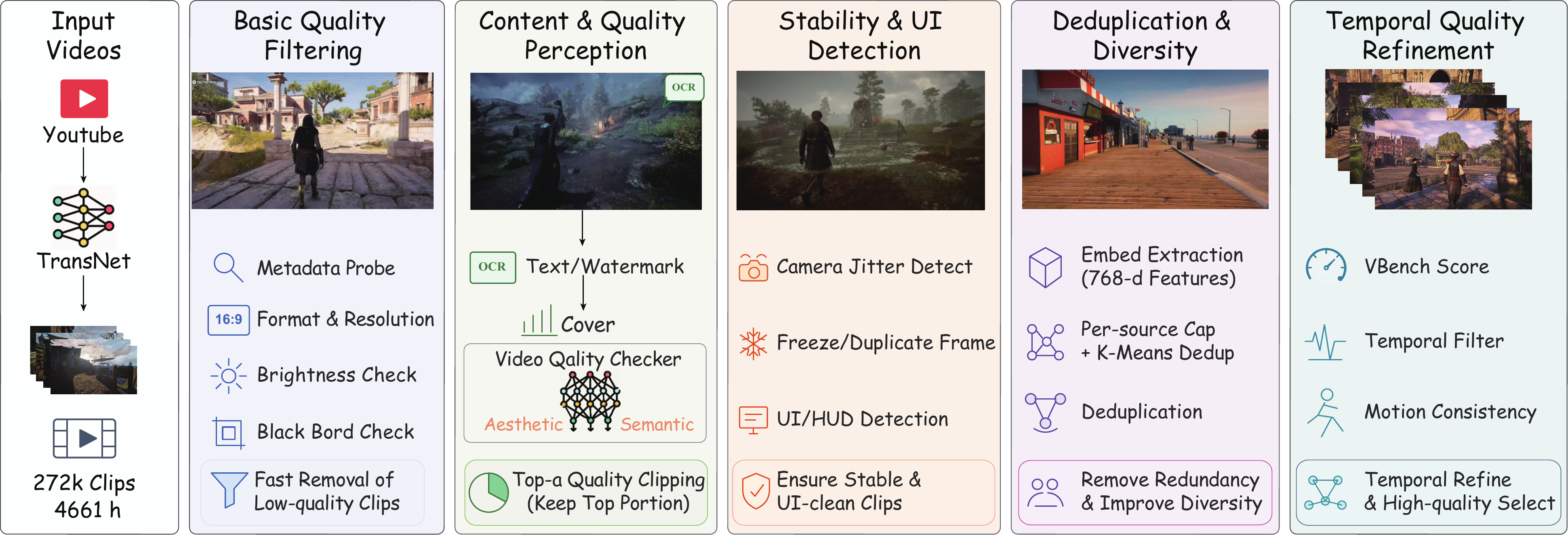}
    \caption{
    Overview of the proposed multi-stage video quality evaluation pipeline. 
    % Starting from segmented gameplay clips, the pipeline performs basic quality filtering, content and quality assessment, stability and UI detection, deduplication and diversity selection, and temporal refinement to remove low-quality or unstable samples while preserving diverse long-duration gameplay dynamics.
    }
    \label{fig:dataset_pipeline}
\vspace{-3mm}
\end{figure}

Existing video datasets for world modeling still exhibit limitations along three axes: fragmented temporal structure caused by short clip segmentation, unstable or absent camera pose estimation, and flat semantic annotations that conflate static context with dynamic events. To address these limitations, we design a unified data construction pipeline that takes a curated pool of long-form gameplay videos and transforms it into structured, minute-scale data units, each consisting of an aligned video segment, a validated camera trajectory, and a multi-granular caption set. The pipeline begins with a \emph{data collection} step (Section~\ref{sec:data_collection}) that builds the raw pool, followed by four processing stages: \emph{scene-consistent segmentation}, \emph{multi-stage video quality evaluation} (illustrated in Figure~\ref{fig:dataset_pipeline}), \emph{camera pose estimation and validation}, and \emph{multi-granular structured captioning}. Each stage targets a specific limitation of prior gameplay datasets.

\subsection{Data Collection}
\label{sec:data_collection}

Constructing a long-horizon gameplay corpus at the scale GameVerse targets requires a source that is simultaneously diverse---across games, perspectives, and behaviors---and openly releasable to the community. Engine recordings used by Genie~\cite{bruce2024genie}, the Matrix-Game family~\cite{zhang2025matrix,he2025matrix,skywork2026matrix3}, Hunyuan-GameCraft~\cite{li2025hunyuan}, and HY-World~1.5~\cite{hyworld2025} satisfy diversity within a single title but are proprietary and typically centered on one or two games. We instead build on YouTube, which is, to our knowledge, the only open source that simultaneously offers long-form gameplay recordings across many AAA and indie titles, both first- and third-person perspectives, and a wide range of player behaviors. All videos are retained at the highest available quality, up to \textbf{1080p at 30--60\,FPS}.

Videos are collected on a per-title basis from a curated set of open-world AAA and indie games chosen to together span both first- and third-person viewpoints. Because the chosen titles already exhibit a wide range of player activities and environments, the resulting source pool naturally covers the eight behavior categories reported in Section~\ref{sec:stats} (\emph{exploring}, \emph{combat}, \emph{walking}, \emph{running}, \emph{driving}, \emph{interacting}, \emph{climbing}, and \emph{idle}) together with diverse environments under varied weather, time-of-day, season, and crowd-density conditions. The collected corpus consists of \textbf{11{,}002 source videos}, totalling \textbf{4{,}692 hours} (195.5 days) and approximately \textbf{986.6\,M frames}, which the four processing stages below convert into the final 188K structured clips.

\subsection{Scene-consistent Segmentation}

Existing datasets typically rely on short clips of a few seconds, which disrupt temporal continuity and limit long-horizon modeling. We start from long-form videos and explicitly construct minute-level continuous segments by removing scene transitions. We first employ TransNetV2~\cite{soucek2024transnet} to detect abrupt scene transitions (hard cuts) based on frame-level transition probabilities. To capture gradual transitions (e.g., camera panning or UI state changes) that are difficult to capture by hard-cut detection alone, we additionally compute a CLIP-based cosine semantic distance between sampled frames, treating large jumps in semantic distance as smooth-transition boundaries. Combining hard-cut detection with semantic-change signals, we extract scene-consistent segments of up to 120 seconds, which can be further concatenated into longer sequences. After segmentation, we obtain \textbf{272,128 clips} totalling \textbf{4,661 hours}, among which 120-second clips contribute approximately \textbf{2,773 hours}.

\subsection{Multi-stage Video Quality Evaluation}

Even after scene-consistent segmentation, raw gameplay videos may still contain subtitles, watermarks, UI overlays, compression artifacts, duplicated content, and unstable temporal dynamics, all of which can introduce distracting visual signals and inconsistent supervision for world modeling. To ensure both visual fidelity and temporal coherence, we design a multi-stage video quality evaluation pipeline (Figure~\ref{fig:dataset_pipeline}) that progressively performs (i) \textit{basic filtering} to remove low-resolution or corrupted videos; (ii) \textit{content and quality assessment} using OCR-based detection to filter subtitles, HUD elements, and other overlays; (iii) \textit{stability and UI detection} to identify unstable motion patterns and temporal artifacts; (iv) \textit{deduplication and diversity selection} to eliminate highly similar samples while preserving diverse gameplay scenarios; and (v) \textit{temporal refinement} using temporal-quality metrics such as COVER~\cite{he2024cover} and VBench~\cite{huang2024vbench} consistency scores. These complementary stages jointly address frame-level visual degradation, temporal inconsistencies, and dataset redundancy. The overall quality of each clip is computed as a weighted sum of its per-stage scores, while stage-specific constraints prevent severe quality defects from being compensated for by high scores in other dimensions. Additionally, we recover a subset of previously filtered clips where the camera remains stationary while scene content continues to evolve, preserving meaningful object and agent dynamics under fixed viewpoints for world-model training. Only clips passing the corresponding quality checks are retained. Detailed thresholds and per-stage criteria are provided in Appendix~\ref{appendix:quality_pipeline}.

\subsection{Camera Pose Estimation and Validation}

Most existing datasets lack reliable camera pose information. Even when available, pose annotations are often noisy or unstable, limiting their usefulness for geometric modeling. To address this, we apply ViPE~\cite{huang2025vipe} to estimate frame-wise camera poses for each video segment, obtaining both intrinsic and extrinsic parameters.

However, pose estimation in dynamic scenes is inherently challenging and often suffers from noise and tracking failures. To ensure trajectory reliability, we validate every estimated pose against three common error modes: \emph{sudden jumps} (large non-physical pose changes between consecutive frames, typically from tracking failures or relocalisation errors), \emph{high-frequency jitter} (local instability producing unnatural oscillations), and \emph{long-term drift} (accumulated errors over time). We score each clip with a unified trajectory-quality function combining local relative-pose increments (capturing jumps and jitter) and a global trajectory smoothness term (capturing drift), and apply threshold-based filtering to remove unreliable samples.

\subsection{Multi-granular Structured Captioning}
\label{sec:pipeline_caption}

Semantic annotations in prior gameplay datasets, such as GameGen-X~\cite{chegamegen} and Hunyuan-GameCraft~\cite{li2025hunyuan}, typically take the form of single-level captions or simple action labels that mix static scene context with dynamic behavior. Even structured annotations often provide limited separation across temporal and semantic scales.

To capture fine-grained gameplay semantics, we introduce a multi-granular structured captioning framework for GameVerse using the vision-language model Kimi-K2.5~\cite{team2026kimi}. As illustrated in Figure~\ref{fig:dataset_gameverse}(e.2), the framework separates static context from dynamic processes and organizes each clip's semantics into four complementary components: \emph{global static}, clip-level environment and scene context (terrain, weather, lighting, and viewpoint); \emph{local static}, relatively stable scene configurations within each temporal segment (functional areas, landmarks, and visible objects); \emph{dynamic events}, changes in the surrounding world (crowd movement, traffic, and environmental transitions); and \emph{agent actions}, the main subject's behavior, described using a unified action vocabulary that separates action semantics from viewpoint. Together, these components provide explicit semantic supervision at multiple granularities, supporting hierarchical and conditional generation. Annotation details and examples are provided in Appendix~\ref{appendix:caption_details}.

\begin{figure}[t]
    \centering
    \includegraphics[width=0.98\linewidth]{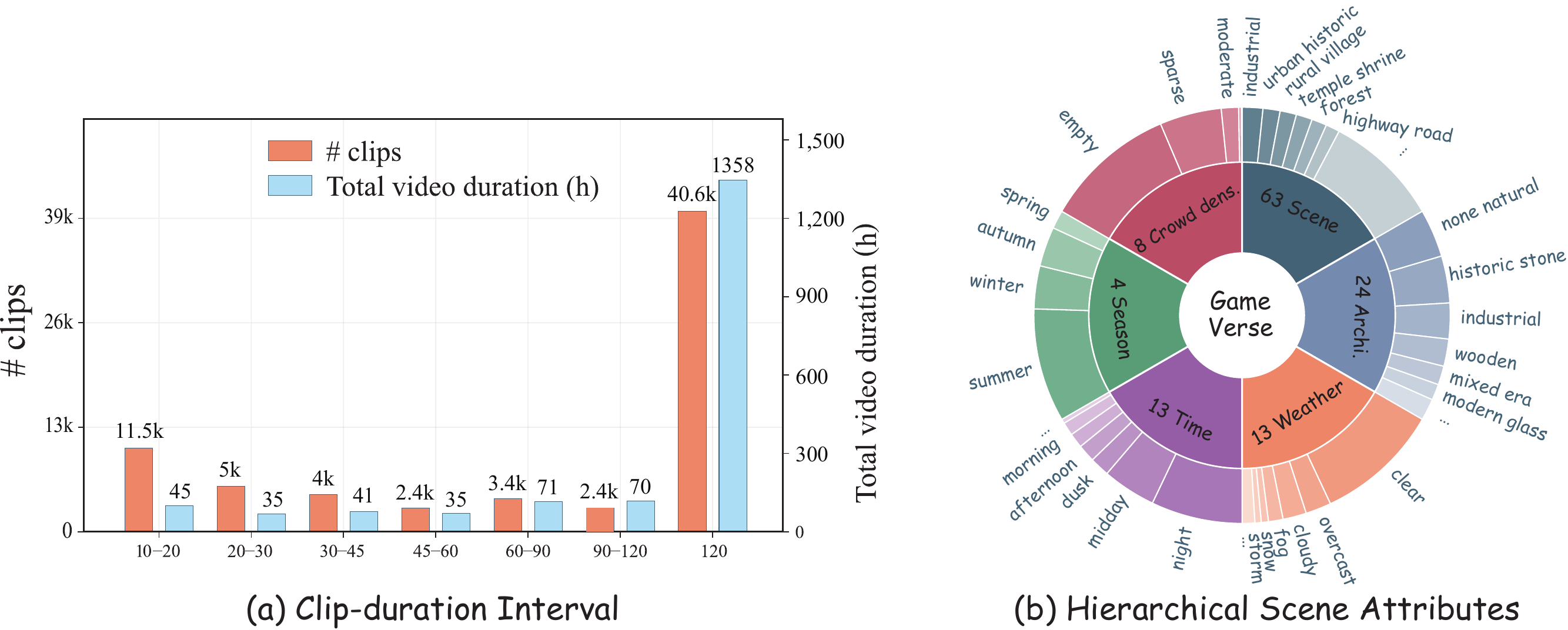}
    \caption{Statistical analysis of \textbf{GameVerse}. (a) Distribution of clip durations and total video hours across different intervals. (b) Hierarchical distribution of scene attributes across six dimensions.
    }
    \label{fig:data_analysis_1}
\vspace{-3mm}
\end{figure}

\section{Dataset Statistics}
\label{sec:stats}

After the construction pipeline of Section~\ref{sec:data_collection}, GameVerse contains \textbf{69.7K high-quality gameplay clips} totalling \textbf{1{,}655 hours}, including \textbf{1{,}358 hours} of two-minute continuous video. We analyse the dataset along four axes that mirror the contribution claims of Section~\ref{sec:intro}: long-horizon temporal scale (Section~\ref{sec:stats_scale}), interaction and environment diversity (Section~\ref{sec:stats_diversity}), camera-trajectory diversity (Section~\ref{sec:stats_camera}), and caption annotation statistics (Section~\ref{sec:stats_caption}).

\subsection{Scale and Long-Horizon Distribution}
\label{sec:stats_scale}

Figure~\ref{fig:data_analysis_1}(a) shows the clip-duration distribution of GameVerse. Unlike prior gameplay video datasets such as GameGen-X~\cite{chegamegen}, Matrix-Game~\cite{zhang2025matrix,he2025matrix,skywork2026matrix3}, and Hunyuan-GameCraft~\cite{li2025hunyuan}, which are dominated by clips shorter than ten seconds, GameVerse preserves substantially longer temporal continuity. Over \textbf{40.6K} clips reach the \textbf{120-second} upper bound and together contribute approximately \textbf{1{,}358 hours} of continuous gameplay video, an order of magnitude longer than the continuous footage exposed in any prior gameplay corpus that we are aware of. Clips of 60\,s or longer account for a further substantial portion of the dataset, providing the kind of minute-scale supervision that is essentially absent from existing gameplay corpora and is required to extend interactive world-model rollouts beyond the one-minute coherence ceiling discussed in Section~\ref{sec:intro}.

\subsection{Interaction and Environment Diversity}
\label{sec:stats_diversity}

GameVerse spans at least \textbf{110 distinct game titles}, with the genre composition shown in panel~(II) of Figure~\ref{fig:dataset_gameverse}. The distribution is long-tailed: the most-represented titles---\emph{God of War}, \emph{Red Dead Redemption~2}, \emph{Ghost of Tsushima}, \emph{Elden Ring}, and \emph{The Witcher~3}---each contribute over 1{,}800 clips, while the tail extends across dozens of additional AAA and indie titles. Across the corpus, approximately \textbf{24\%} of clips are first-person, \textbf{73\%} are third-person follow camera, and \textbf{3\%} use specialised perspectives (drone, vehicle-mounted, gimbal-follow, stationary). This 1:3 first-to-third-person ratio supplies substantial supervision for both FPV and TPV settings, in sharp contrast to the Minecraft-only Matrix-Game~\cite{zhang2025matrix,he2025matrix} and to Sekai~\cite{lisekai}, which is restricted to first-person walking footage. Figure~\ref{fig:dataset_gameverse} also presents the distribution of player-behavior categories---\emph{exploring}, \emph{combat}, \emph{walking}, \emph{driving}, \emph{interacting}, \emph{climbing}, \emph{running}, and \emph{idle}---spanning basic locomotion to complex object interaction and high-speed vehicle control; exploring, combat, walking, and driving together contribute the largest share of total video duration. Figure~\ref{fig:data_analysis_1}(b) further shows the hierarchical scene-attribute distribution: the corpus spans urban, rural, forest, highway, temple, and industrial environments under varied architectural styles, weather conditions, time-of-day, seasons, and crowd-density levels.

\begin{figure}[t]
    \centering
    \includegraphics[width=0.9\linewidth]{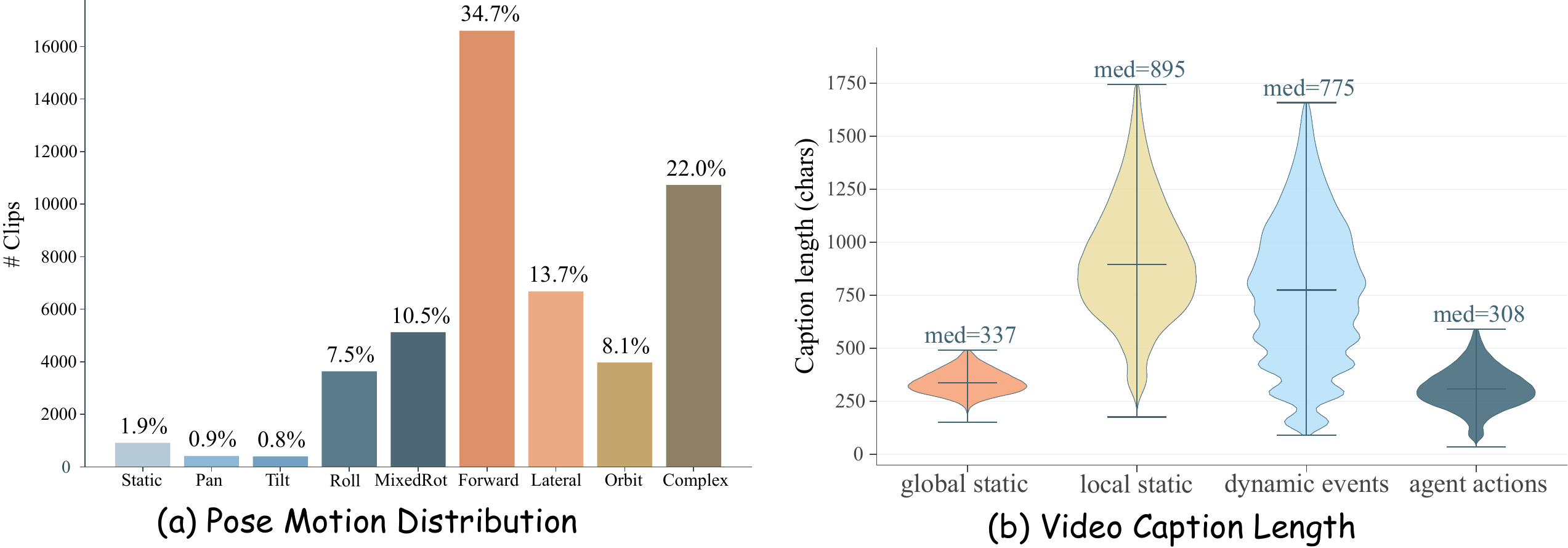}
    \caption{Analysis of camera motion and semantic annotations in \textbf{GameVerse}. (a) Distribution of camera motion patterns, including static, rotational, translational, and complex motions. (b) Distribution of caption lengths across different semantic levels, including global static context, local static structure, dynamic events, and agent actions.
    }
    \label{fig:data_analysis_2}
\vspace{-3mm}
\end{figure}

\subsection{Camera Trajectory Diversity}
\label{sec:stats_camera}

Figure~\ref{fig:data_analysis_2}(a) shows the distribution of camera-motion patterns in GameVerse. The corpus covers \emph{static}, \emph{rotational}, \emph{translational}, and \emph{complex} motions. Forward motion accounts for the largest share (\textbf{34.7\%}), reflecting the prevalence of continuous navigation and exploration in gameplay. Complex motions that couple translation with rotation contribute a further \textbf{22.0\%}; these clips are particularly challenging for long-horizon world modeling because they require simultaneous geometric and identity consistency under non-trivial viewpoint change. Lateral movement, orbit motion, and mixed rotational patterns are also well represented, indicating substantial diversity along the trajectory axis. By contrast, real-world walking-video datasets such as Sekai are typically dominated by forward translation with limited rotational complexity, and gameplay datasets that obtain engine-level poses are restricted to a single title's controller scheme. The richer camera dynamics in GameVerse therefore enable supervision for geometry-aware generation, camera-controllable video synthesis, and interactive world modeling under genuine viewpoint change.

\subsection{Caption Annotation Statistics}
\label{sec:stats_caption}

\begin{wrapfigure}{r}{0.46\textwidth}
    \centering
    \vspace{-0.4 cm}
    \includegraphics[width=1.0\linewidth]{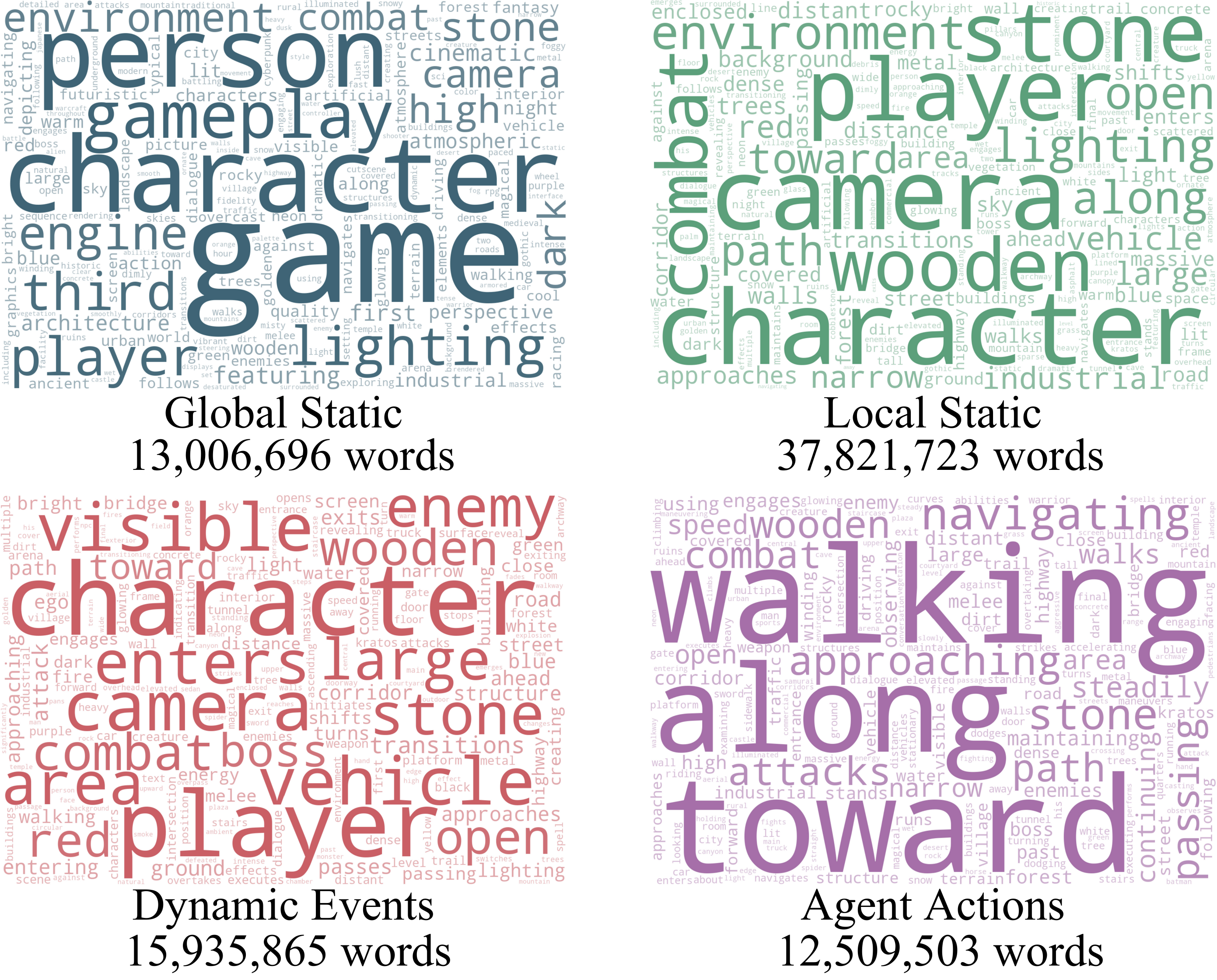}
    \vspace{-0.4 cm}
    \caption{ Word-cloud visualization of the four semantic caption components in GameVerse. 
    % The distributions highlight the semantic diversity and distinct focus of different annotation levels.
    }
    \label{fig:caption_wordcloud}
    \vspace{-0.4 cm}
\end{wrapfigure}

Figure~\ref{fig:data_analysis_2}(b) shows the caption-length distributions across the four semantic levels introduced in Section~\ref{sec:pipeline_caption}. \emph{Local static} descriptions and \emph{dynamic events} are substantially longer, with median lengths of \textbf{895} and \textbf{775} characters respectively, reflecting the rich environmental structure and dynamic scene changes in gameplay video. \emph{Global static} and \emph{agent action} captions are more concise: global-static captions describe high-level scene context, and agent actions are drawn from a constrained vocabulary so that the controllable channel can be modulated cleanly during conditional generation. Compared to the single-line whole-clip captions of Sekai and the clip-level captions of GameGen-X, the four-level schema produces substantially more total semantic information per clip while remaining structured enough to act as a hierarchical conditioning signal.

Figure~\ref{fig:caption_wordcloud} shows that the four levels concentrate on largely disjoint vocabularies---scene attributes, spatial layouts, environment transitions, and explicit motion---confirming that the schema produces semantically separated annotations rather than overlapping rephrasings of the same clip.

\section{Experiments}
\label{sec:experiments}

\subsection{Experimental Setup}
We evaluate 60-second image-to-video generation on the same 103 inputs using LingBot~\cite{robbyant2026lingbot}, HY-World~1.5~\cite{hyworld2025,sun2025worldplay}, HY-World-FT, and LTX-FT. \textbf{HY-World-FT} is obtained by fine-tuning HY-World~1.5 on approximately 600 hours of GameVerse for 3{,}000 steps, using the step-3{,}000 checkpoint without changing the architecture. \textbf{LTX-FT} is a GameVerse-adapted LTX-Video~2.3 model~\cite{lightricks2026ltx2}. The fixed-backbone comparison measures the benefit of additional GameVerse fine-tuning under a shared evaluation protocol; comparisons across model families characterize complete systems with different architectures and control interfaces.

\subsection{Evaluation Metrics}
We evaluate generation performance from two complementary perspectives. \emph{\textbf{Video generation quality}} is assessed using VBench~\cite{huang2024vbench} across six dimensions: Subject Consistency, Background Consistency, Motion Smoothness, Aesthetic Quality, Imaging Quality, and Overall Consistency, collectively measuring temporal coherence, motion stability, visual fidelity, and text--video alignment. \emph{\textbf{World generation quality and controllability}} are evaluated using WorldScore~\cite{duan2025worldscore}, covering camera and object control, content alignment, 3D consistency, photometric consistency, style consistency, and subjective quality. We report all seven normalized WorldScore components alongside their arithmetic mean (WS-Static). Higher scores indicate better performance across all reported metrics. Detailed definitions and evaluation protocols are provided in Appendix~\ref{appendix:eval_metrics}.

\begin{table}[t]
\centering
\small
\setlength{\tabcolsep}{4pt}
\caption{Overall VBench evaluation results on the full evaluation set. $\uparrow$ indicates higher is better. \textbf{Bold} indicates the best reported value in each column.}
\label{tab:vbench_60s}
\begin{adjustbox}{max width=\linewidth}
\begin{tabular}{lcccccc}
\toprule
Model & Subj. Cons.$\uparrow$ & Bkg. Cons.$\uparrow$ & Mot. Smth.$\uparrow$ & Aes. Qual.$\uparrow$ & Img. Qual.$\uparrow$ & Overall Cons.$\uparrow$ \\
\midrule
LingBot & 74.30 & 88.94 & 97.36 & \textbf{6.14} & \textbf{56.70} & \textbf{18.25} \\
HY-World 1.5 & 66.92 & 81.77 & 96.26 & 4.29 & 48.99 & 14.48 \\
HY-World-FT & 68.45 & 83.30 & 96.81 & 4.65 & 49.34 & 15.69 \\
LTX-FT & \textbf{79.17} & \textbf{90.63} & \textbf{99.53} & 5.79 & 49.36 & 17.57 \\
\bottomrule
\end{tabular}
\end{adjustbox}
\vspace{-5pt}
\end{table}

\begin{table}[t]
\centering
\small
\setlength{\tabcolsep}{4pt}
\caption{Overall WorldScore evaluation results on the full evaluation set. $\uparrow$ indicates higher is better. Best results are highlighted in \textbf{bold}.}
\label{tab:worldscore_60s}
\begin{adjustbox}{max width=\linewidth}
\begin{tabular}{lcccccccc}
\toprule
Model & WS-Static$\uparrow$ & Cam. Ctrl.$\uparrow$ & Obj. Ctrl.$\uparrow$ & Content$\uparrow$ & 3D Cons.$\uparrow$ & Photo.$\uparrow$ & Style$\uparrow$ & Subj. Qual.$\uparrow$ \\
\midrule
LingBot & 46.07 & \textbf{31.69} & 45.56 & \textbf{50.18} & 63.25 & 22.71 & 86.06 & \textbf{23.05} \\
HY-World 1.5 & 53.33 & 28.58 & 45.54 & 48.16 & 66.94 & 83.15 & 87.52 & 13.42 \\
HY-World-FT & 55.13 & 28.90 & \textbf{46.26} & 49.59 & 73.42 & 84.08 & 88.11 & 15.58 \\
LTX-FT & \textbf{59.65} & 24.33 & 44.93 & 48.47 & \textbf{95.13} & \textbf{98.45} & \textbf{90.04} & 16.22 \\
\bottomrule
\end{tabular}
\end{adjustbox}
\vspace{-5pt}
\end{table}

\subsection{Minute-Scale Generation Performance}

\paragraph{VBench Evaluation.}
As shown in Table~\ref{tab:vbench_60s}, LTX-FT achieves the best temporal consistency across 60-second rollouts, leading in Subject Consistency (79.17), Background Consistency (90.63), and Motion Smoothness (99.53). Meanwhile, LingBot performs better on perceptual quality and text--video alignment, achieving the highest Aesthetic Quality (6.14), Imaging Quality (56.70), and Overall Consistency (18.25). Notably, under the same HY-World~1.5 backbone, fine-tuning on GameVerse consistently improves all six VBench metrics. In particular, Subject Consistency increases from 66.92 to 68.45, Background Consistency from 81.77 to 83.30, and Overall Consistency from 14.48 to 15.69, demonstrating the effectiveness of GameVerse in enhancing temporal coherence and generation quality without architectural modifications.

\paragraph{WorldScore Evaluation.}
Table~\ref{tab:worldscore_60s} evaluates world generation quality and controllability. LTX-FT achieves the highest WS-Static score (59.65), leading in 3D Consistency (95.13), Photometric Consistency (98.45), and Style Consistency (90.04). LingBot performs best in Camera Control, Content Alignment, and Subjective Quality, while HY-World-FT leads in Object Control. LingBot's stronger camera control may partly benefit from ground-truth pose supervision and dedicated conditioning mechanisms, whereas GameVerse relies on ViPE-estimated trajectories (see Appendix~\ref{appendix:pose_sources}). Importantly, fine-tuning the same HY-World backbone (\textbf{Hy-World-FT}) on GameVerse improves all seven WorldScore components, increasing WS-Static from 53.33 to 55.13, 3D Consistency from 66.94 to 73.42 (+6.48), and Camera Control from 28.58 to 28.90. These results demonstrate the effectiveness of GameVerse in improving geometric consistency and controllability for minute-scale generation.

\subsection{Long-Horizon Consistency Analysis}

\paragraph{Reference-frame Similarity.}
To further assess long-horizon stability, we sample generated frames at 2\,Hz and compute their CLIP ViT-B/32~\cite{radford2021learning} similarity to the input frame. As shown in Table~\ref{tab:temporal_activity}(a), LTX-FT achieves the highest similarity at 30 seconds (0.785) and the latest reported threshold crossing (50.7s). Notably, under the same HY-World backbone, GameVerse fine-tuning improves the 30-second similarity from 0.609 to 0.642 and delays the first similarity drop below 0.6 from 30.4s to 37.9s. These results suggest improved reference preservation and reduced visual drift over extended rollouts, although viewpoint changes may also affect similarity.

\paragraph{Performance across Gameplay Activities.}
Table~\ref{tab:temporal_activity}(b) further evaluates Subject Consistency across four gameplay activities. LTX-FT achieves the highest scores across driving, riding, running, and walking. Under the same backbone, HY-World-FT consistently outperforms HY-World~1.5, with improvements of 1.15, 1.85, 2.49, and 1.58 points, respectively. These consistent gains across diverse gameplay activities further support the effectiveness of GameVerse for long-horizon generation.

\begin{table*}[t]
\centering
\small
\setlength{\tabcolsep}{4pt}
\caption{Temporal stability and activity-wise Subject Consistency of 60s generations. $\uparrow$ indicates higher is better. Best results are highlighted in bold.}
\label{tab:temporal_activity}

\begin{minipage}[t]{0.424\textwidth}
\centering
\textbf{(a) Reference-frame Similarity}
\vspace{4pt}

\begin{adjustbox}{max width=\linewidth}
\begin{tabular}{lcc}
\toprule
Model & First-drop (s)$\uparrow$ & CLIP@30s$\uparrow$ \\
\midrule
LingBot & 49.0 & 0.680 \\
HY-World 1.5 & 30.4 & 0.609 \\
HY-World-FT & 37.9 & 0.642 \\
LTX-FT & \textbf{50.7} & \textbf{0.785} \\
\bottomrule
\end{tabular}
\end{adjustbox}
\end{minipage}
\hfill
\begin{minipage}[t]{0.55\textwidth}
\centering
\textbf{(b) Subject Consistency}
\vspace{4pt}

\begin{adjustbox}{max width=\linewidth}
\begin{tabular}{lcccc}
\toprule
Model & Driving$\uparrow$ & Riding$\uparrow$ & Running$\uparrow$ & Walking$\uparrow$ \\
\midrule
LingBot & 73.23 & 76.37 & 74.01 & 73.69 \\
HY-World 1.5 & 65.02 & 63.47 & 66.05 & 71.48 \\
HY-World-FT & 66.17 & 65.32 & 68.54 & 73.06 \\
LTX-FT & \textbf{81.44} & \textbf{79.13} & \textbf{76.16} & \textbf{80.74} \\
\bottomrule
\end{tabular}
\end{adjustbox}
\end{minipage}
\vspace{-5pt}

\end{table*}

\subsection{Annotation Quality Validation}

\paragraph{Camera trajectories.}
We assess trajectory completeness and temporal stability after quality filtering, and compare ViPE estimates with an independent DROID-SLAM estimator~\cite{teed2021droid}. After Sim(3) alignment, rotation RPE ranges from 0.75 to 1.47 degrees and translation-direction error from 4.84 to 11.03 degrees across six motion categories. The audited retained trajectories have complete valid poses; mixed rotation exhibits the largest cross-estimator errors. These measurements characterize consistency between two estimators, rather than absolute pose accuracy against ground truth. Full motion-wise results are provided in Appendix~\ref{appendix:pose_quality}.

% \begin{wraptable}{r}{0.48\linewidth}
% \vspace{-2pt}
% \raggedleft
% \small
% \setlength{\tabcolsep}{4pt}
% \renewcommand{\arraystretch}{1.12}

% \begin{threeparttable}
% \caption{Human evaluation of structured captions on 400 gameplay clips.}
% \label{tab:caption_human}
% \vspace{3pt}

% \begin{tabular}{@{}l@{\hspace{12pt}}r@{}}
% \toprule
% Human-audit Measure & Result \\
% \midrule
% Scene/event accuracy & 4.82 / 5 \\
% Agent-action accuracy & 4.66 / 5 \\
% Temporal alignment & 4.61 / 5 \\
% Camera-motion accuracy & 4.55 / 5 \\
% Overall alignment & 4.64 / 5 \\
% Clip-level hallucination & 5.5\% \\
% Temporal mismatch & 4.0\% \\
% Krippendorff's $\alpha$ & 0.76 \\
% \bottomrule
% \end{tabular}
% \end{threeparttable}
% \vspace{-6pt}
% \end{wraptable}

\begin{wraptable}{r}{0.36\linewidth}
\vspace{-2pt}
\centering
\small
\setlength{\tabcolsep}{4pt}
\renewcommand{\arraystretch}{1.12}

\begin{threeparttable}
\caption{Human evaluation of structured captions on 400 clips.}
\label{tab:caption_human}
\vspace{3pt}

\begin{tabularx}{\linewidth}{@{}>{\raggedright\arraybackslash}X r@{}}
\toprule
Human-audit Measure & Result \\
\midrule
Scene/event accuracy & 4.82 / 5 \\
Agent-action accuracy & 4.66 / 5 \\
Temporal alignment & 4.61 / 5 \\
Camera-motion accuracy & 4.55 / 5 \\
Overall alignment & 4.64 / 5 \\
Clip-level hallucination & 5.5\% \\
Temporal mismatch & 4.0\% \\
Krippendorff's $\alpha$ & 0.76 \\
\bottomrule
\end{tabularx}

\end{threeparttable}
\vspace{-6pt}
\end{wraptable}

\paragraph{Structured Captions.}
To assess the quality of our multi-granular semantic annotations, we conduct a video-grounded human evaluation on 400 clips stratified by game, viewpoint, motion type, and duration. Three independent annotators evaluate the accuracy and temporal alignment of each caption against its corresponding video. As shown in Table~\ref{tab:caption_human}, the annotations achieve high scores in scene/event accuracy (4.82/5) and agent-action accuracy (4.66/5), while temporal alignment (4.61/5) and camera-motion accuracy (4.55/5) also show favorable results. The overall alignment score reaches 4.64/5, with consistent judgments among annotators (Krippendorff's $\alpha=0.76$). Furthermore, the clip-level hallucination and temporal-mismatch rates are limited to 5.5\% and 4.0\%, respectively. These results demonstrate the effectiveness of our structured captioning pipeline in capturing fine-grained gameplay semantics, while highlighting remaining challenges in annotation accuracy and temporal alignment.

\paragraph{Case Study.}
We provide representative gameplay visualizations in Appendix~\ref{appendix:visualization} and multi-granular structured annotation examples in Appendix~\ref{appendix:caption_details}, illustrating the diversity of environments, continuous scene evolution, and fine-grained semantic supervision in GameVerse. 
We further present qualitative generation results of LTX-Video~2.3 fine-tuned on GameVerse in Appendix~\ref{appendix:case_study}. 
Conditioned on an initial frame, camera trajectories, and structured semantic captions, the model generates temporally coherent gameplay sequences across diverse scenarios, including outdoor exploration, urban navigation, horseback riding, and nighttime driving. 
The results exhibit consistent scene structures, smooth camera-aware motion, and coherent environmental evolution. 
Additional camera-controlled examples demonstrate the model's ability to follow complex trajectories while maintaining spatial and temporal consistency, highlighting the potential of GameVerse for long-horizon controllable world generation.

\section{Conclusion and Limitations}

We present \textbf{GameVerse}, a minute-scale gameplay dataset for long-horizon interactive world modeling, comprising 69.7K clips totaling \textbf{1{,}655 hours}, including \textbf{1{,}358 hours of continuous two-minute} gameplay. Built through a unified pipeline, GameVerse provides quality-filtered frame-level camera trajectories and multi-granular structured captions, jointly capturing long-term scene evolution, camera motion, and fine-grained interaction dynamics. Experiments on 60-second generation demonstrate improvements in temporal coherence, geometric consistency, and visual quality, with same-backbone comparisons further validating the benefits of GameVerse fine-tuning. Additional annotation-quality assessments support the reliability of its semantic and geometric supervision. We envision GameVerse as a valuable resource for advancing long-horizon interactive world modeling and enabling more coherent, controllable, and persistent virtual worlds.

\paragraph{Limitations.}
GameVerse focuses on gameplay environments, with cross-domain generalization to real-world scenarios remaining unexplored. Despite multi-stage validation, estimated camera trajectories and generated captions may retain residual errors. Continuous clips are limited to two minutes, while generation experiments cover up to 60 seconds, leaving longer rollouts and closed-loop interaction for future work. Moreover, the individual contributions of clip duration, camera supervision, and structured captions remain to be isolated, and large-scale data construction is computationally intensive.

\begin{ack}
This work was conducted during an internship at Alaya Lab. We sincerely thank Kaipeng Zhang and Yongtao Ge, the project leader, for their valuable guidance and continuous support throughout this research. We are also grateful to our colleagues and collaborators for their insightful discussions, constructive feedback, and contributions to the development of GameVerse.

\textbf{Author Contributions.} Kang He led the overall research, including the design and implementation of the GameVerse dataset construction pipeline, model training, experimental evaluation, and manuscript preparation. Wenshuo Peng contributed substantially to the large-scale camera-pose and structured caption annotation of the dataset.

\textbf{Funding and Competing Interests.} This research was primarily supported by Alaya Lab through the provision of research data resources. Alaya Lab has also authorized the public release of the resulting dataset metadata and derived annotations. The authors declare no competing interests.
\end{ack}

% \begin{ack}
% Use unnumbered first level headings for the acknowledgments. All acknowledgments

% Do {\bf not} include this section in the anonymized submission, only in the final paper. You can use the \texttt{ack} environment provided in the style file to automatically hide this section in the anonymized submission.
% \end{ack}

\bibliographystyle{plain}
\bibliography{neurips_2026}

@article{decart2024oasis,
  title={Oasis: A universe in a transformer},
  author={Decart, Etched and McIntyre, Quinn and Campbell, Spruce and Chen, Xinlei and Wachen, Robert},
  journal={URL: https://oasis-model. github. io},
  volume={2},
  number={3},
  pages={6},
  year={2024}
}

@article{zhang2025matrix,
  title={Matrix-game: Interactive world foundation model},
  author={Zhang, Yifan and Peng, Chunli and Wang, Boyang and Wang, Puyi and Zhu, Qingcheng and Kang, Fei and Jiang, Biao and Gao, Zedong and Li, Eric and Liu, Yang and others},
  journal={arXiv preprint arXiv:2506.18701},
  year={2025}
}

@article{he2025matrix,
  title={Matrix-game 2.0: An open-source real-time and streaming interactive world model},
  author={He, Xianglong and Peng, Chunli and Liu, Zexiang and Wang, Boyang and Zhang, Yifan and Cui, Qi and Kang, Fei and Jiang, Biao and An, Mengyin and Ren, Yangyang and others},
  journal={arXiv preprint arXiv:2508.13009},
  year={2025}
}

@article{team2026kimi,
  title={Kimi K2. 5: Visual Agentic Intelligence},
  author={Team, Kimi and Bai, Tongtong and Bai, Yifan and Bao, Yiping and Cai, SH and Cao, Yuan and Charles, Y and Che, HS and Chen, Cheng and Chen, Guanduo and others},
  journal={arXiv preprint arXiv:2602.02276},
  year={2026}
}

@article{ju2024miradata,
  title={Miradata: A large-scale video dataset with long durations and structured captions},
  author={Ju, Xuan and Gao, Yiming and Zhang, Zhaoyang and Yuan, Ziyang and Wang, Xintao and Zeng, Ailing and Xiong, Yu and Xu, Qiang and Shan, Ying},
  journal={Advances in Neural Information Processing Systems},
  volume={37},
  pages={48955--48970},
  year={2024}
}

@article{skywork2026matrix3,
  title={Matrix-Game 3.0: Real-Time and Streaming Interactive World Model with Long-Horizon Memory},
  author={{Skywork AI Matrix-Game Team}},
  journal={arXiv preprint arXiv:2604.08995},
  year={2026}
}

@article{li2025hunyuan,
  title={Hunyuan-gamecraft: High-dynamic interactive game video generation with hybrid history condition},
  author={Li, Jiaqi and Tang, Junshu and Xu, Zhiyong and Wu, Longhuang and Zhou, Yuan and Shao, Shuai and Yu, Tianbao and Cao, Zhiguo and Lu, Qinglin},
  journal={arXiv preprint arXiv:2506.17201},
  volume={2},
  number={3},
  pages={6},
  year={2025}
}

@inproceedings{li2025vmem,
  title={Vmem: Consistent interactive video scene generation with surfel-indexed view memory},
  author={Li, Runjia and Torr, Philip and Vedaldi, Andrea and Jakab, Tomas},
  booktitle={Proceedings of the IEEE/CVF International Conference on Computer Vision},
  pages={25690--25699},
  year={2025}
}

@article{sun2025worldplay,
  title={Worldplay: Towards long-term geometric consistency for real-time interactive world modeling},
  author={Sun, Wenqiang and Zhang, Haiyu and Wang, Haoyuan and Wu, Junta and Wang, Zehan and Wang, Zhenwei and Wang, Yunhong and Zhang, Jun and Wang, Tengfei and Guo, Chunchao},
  journal={arXiv preprint arXiv:2512.14614},
  year={2025}
}

@article{hyworld2025,
  title={HY-World 1.5: A Systematic Framework for Interactive World Modeling with Real-Time Latency and Geometric Consistency},
  author={Team HunyuanWorld},
  journal={arXiv preprint},
  year={2025}
}

@inproceedings{bruce2024genie,
  author={Bruce, Jake and Dennis, Michael D and Edwards, Ashley and Parker-Holder, Jack and Shi, Yuge and Hughes, Edward and Lai, Matthew and Mavalankar, Aditi and Steigerwald, Richie and Apps, Chris and others},
  title={Genie: Generative Interactive Environments},
  booktitle={Proceedings of the International Conference on Machine Learning},
  pages={4603--4623},
  year={2024}
}

@article{he2026sekai2,
  title={Sekai2: From World Exploration to Interactive World Modeling},
  author={He, Kang and Peng, Wenshuo and Gao, Zihui and Tan, Jiaming and Zhang, Kaipeng and Ge, Yongtao},
  journal={arXiv preprint arXiv:2608.09449},
  year={2026}
}

@article{xu2026worldmark,
  title={WorldMark: A unified benchmark suite for interactive video world models},
  author={Xu, Xiaojie and Lin, Zhengyuan and He, Kang and Feng, Yukang and Mao, Xiaofeng and Yin, Yuanyang and Ge, Yongtao and Zhang, Kaipeng},
  journal={arXiv preprint arXiv:2604.21686},
  year={2026}
}

@article{team2026alayaworld,
  title={AlayaWorld: Interactive Long-Horizon World Modeling--Full Technical Report},
  author={Team, AlayaWorld and Zhang, Kaipeng and Li, Chuanhao and Zhan, Yifan and Ge, Yongtao and Yin, Yuanyang and Tan, Jiaming and He, Kang and Fan, Liaoyuan and Zhai, Mingliang and others},
  journal={arXiv preprint arXiv:2607.18367},
  year={2026}
}

@inproceedings{duan2025worldscore,
  title={Worldscore: A unified evaluation benchmark for world generation},
  author={Duan, Haoyi and Yu, Hong-Xing and Chen, Sirui and Fei-Fei, Li and Wu, Jiajun},
  booktitle={Proceedings of the IEEE/CVF international conference on computer vision},
  pages={27713--27724},
  year={2025}
}

@article{yin2026alaya,
  title={Alaya-EVOKE: From linear-scaling supervision to endless world},
  author={Yin, Yuanyang and Wang, Gongxuan and Zhan, Yifan and Li, Chuanhao and Zhang, Kaipeng and Zhao, Feng},
  journal={arXiv preprint arXiv:2608.13546},
  year={2026}
}

@inproceedings{radford2021learning,
  title={Learning transferable visual models from natural language supervision},
  author={Radford, Alec and Kim, Jong Wook and Hallacy, Chris and Ramesh, Aditya and Goh, Gabriel and Agarwal, Sandhini and Sastry, Girish and Askell, Amanda and Mishkin, Pamela and Clark, Jack and others},
  booktitle={International conference on machine learning},
  pages={8748--8763},
  year={2021},
}

@article{teed2021droid,
  title={Droid-slam: Deep visual slam for monocular, stereo, and rgb-d cameras},
  author={Teed, Zachary and Deng, Jia},
  journal={Advances in neural information processing systems},
  volume={34},
  pages={16558--16569},
  year={2021}
}

@article{wan2025wan,
  title={Wan: Open and advanced large-scale video generative models},
  author={Wan, Team and Wang, Ang and Ai, Baole and Wen, Bin and Mao, Chaojie and Xie, Chen-Wei and Chen, Di and Yu, Feiwu and Zhao, Haiming and Yang, Jianxiao and others},
  journal={arXiv preprint arXiv:2503.20314},
  year={2025}
}

@inproceedings{DBLP:conf/iclr/ValevskiLAF25,
  author       = {Dani Valevski and
                  Yaniv Leviathan and
                  Moab Arar and
                  Shlomi Fruchter},
  title        = {Diffusion Models Are Real-Time Game Engines},
  booktitle    = {Proceedings of the thirteenth International Conference on Learning Representations,
                  {ICLR} 2025, Singapore, April 24-28, 2025},
  year         = {2025}
}

@inproceedings{he2024cover,
  title={Cover: A comprehensive video quality evaluator},
  author={He, Chenlong and Zheng, Qi and Zhu, Ruoxi and Zeng, Xiaoyang and Fan, Yibo and Tu, Zhengzhong},
  booktitle={Proceedings of the IEEE/CVF conference on computer vision and pattern recognition},
  pages={5799--5809},
  year={2024}
}

@inproceedings{soucek2024transnet,
  title={Transnet v2: An effective deep network architecture for fast shot transition detection},
  author={Soucek, Tom{\'a}s and Lokoc, Jakub},
  booktitle={Proceedings of the 32nd ACM International Conference on Multimedia},
  pages={11218--11221},
  year={2024}
}

@inproceedings{xiaoworldmem,
  title={WorldMem: Long-term Consistent World Simulation with Memory},
  author={Xiao, Zeqi and Yushi, LAN and Zhou, Yifan and Ouyang, Wenqi and Yang, Shuai and Zeng, Yanhong and Pan, Xingang},
  booktitle={Proceedings of the Thirty-ninth Annual Conference on Neural Information Processing Systems},
  year={2025}
}

@inproceedings{chegamegen,
  title={GameGen-X: Interactive Open-world Game Video Generation},
  author={Che, Haoxuan and He, Xuanhua and Liu, Quande and Jin, Cheng and Chen, Hao},
  booktitle={Proceedings of the Thirteenth International Conference on Learning Representations},
  year={2025}
}

@inproceedings{krishna2017dense,
  title={Dense-captioning events in videos},
  author={Krishna, Ranjay and Hata, Kenji and Ren, Frederic and Fei-Fei, Li and {Carlos Niebles}, Juan},
  booktitle={Proceedings of the IEEE international conference on computer vision},
  pages={706--715},
  year={2017}
}

@inproceedings{wanginternvid,
  title={InternVid: A Large-scale Video-Text Dataset for Multimodal Understanding and Generation},
  author={Wang, Yi and He, Yinan and Li, Yizhuo and Li, Kunchang and Yu, Jiashuo and Ma, Xin and Li, Xinhao and Chen, Guo and Chen, Xinyuan and Wang, Yaohui and others},
  booktitle={Proceedings of the Twelfth International Conference on Learning Representations},
  year={2024}
}

@inproceedings{lisekai,
  title={Sekai: A Video Dataset towards World Exploration},
  author={Li, Zhen and Li, Chuanhao and Mao, Xiaofeng and Lin, Shaoheng and Li, Ming and Zhao, Shitian and Li, Xinyue and Feng, Yukang and Sun, Jianwen and Li, Zizhen and others},
  booktitle={Proceedings of the Thirty-ninth Annual Conference on Neural Information Processing Systems Datasets and Benchmarks Track},
  year={2025}
}

@article{he2024cameractrl,
  title={Cameractrl: Enabling camera control for text-to-video generation},
  author={He, Hao and Xu, Yinghao and Guo, Yuwei and Wetzstein, Gordon and Dai, Bo and Li, Hongsheng and Yang, Ceyuan},
  journal={arXiv preprint arXiv:2404.02101},
  year={2024}
}

@inproceedings{wang2024motionctrl,
  title={Motionctrl: A unified and flexible motion controller for video generation},
  author={Wang, Zhouxia and Yuan, Ziyang and Wang, Xintao and Li, Yaowei and Chen, Tianshui and Xia, Menghan and Luo, Ping and Shan, Ying},
  booktitle={Proceedings of the ACM SIGGRAPH 2024 Conference Papers},
  pages={1--11},
  year={2024}
}

@article{huang2025vipe,
  title={Vipe: Video pose engine for 3d geometric perception},
  author={Huang, Jiahui and Zhou, Qunjie and Rabeti, Hesam and Korovko, Aleksandr and Ling, Huan and Ren, Xuanchi and Shen, Tianchang and Gao, Jun and Slepichev, Dmitry and Lin, Chen-Hsuan and others},
  journal={arXiv preprint arXiv:2508.10934},
  year={2025}
}

@inproceedings{huang2024vbench,
  title={VBench: Comprehensive Benchmark Suite for Video Generative Models},
  author={Huang, Ziqi and He, Yinan and Yu, Jiashuo and Zhang, Fan and Si, Chenyang and Jiang, Yuming and Zhang, Yuanhan and Wu, Tianxing and Jin, Qingyang and Chanpaisit, Nattapol and Wang, Yaohui and Chen, Xinyuan and Wang, Limin and Lin, Dahua and Qiao, Yu and Liu, Ziwei},
  booktitle={Proceedings of the IEEE/CVF Conference on Computer Vision and Pattern Recognition},
  year={2024}
}

@article{lightricks2026ltx2,
  title={{LTX-2}: Efficient Joint Audio-Visual Foundation Model},
  author={{Lightricks}},
  journal={arXiv preprint arXiv:2601.03233},
  year={2026}
}

@article{robbyant2026lingbot,
  title={Advancing Open-source World Models},
  author={Gao, Zelin and Wang, Qiuyu and Zeng, Yanhong and Zhu, Jiapeng and Cheng, Ka Leong and Li, Yixuan and Wang, Hanlin and Xu, Yinghao and Ma, Shuailei and Chen, Yihang and Liu, Jie and Cheng, Yansong and Yao, Yao and Zhu, Jiayi and Meng, Yihao and Zheng, Kecheng and Bai, Qingyan and Chen, Jingye and Shen, Zehong and Yu, Yue and Zhu, Xing and Shen, Yujun and Ouyang, Hao},
  journal={arXiv preprint arXiv:2601.20540},
  year={2026}
}

@inproceedings{miech2019howto100m,
  title={HowTo100M: Learning a Text-Video Embedding by Watching Hundred Million Narrated Video Clips},
  author={Miech, Antoine and Zhukov, Dimitri and Alayrac, Jean-Baptiste and Tapaswi, Makarand and Laptev, Ivan and Sivic, Josef},
  booktitle={Proceedings of the 2019 IEEE/CVF International Conference on Computer Vision (ICCV)},
  pages={2630--2640},
  year={2019}
}

@inproceedings{DBLP:conf/cvpr/HeilbronEGN15,
  author       = {Fabian Caba Heilbron and
                  Victor Escorcia and
                  Bernard Ghanem and
                  Juan Carlos Niebles},
  title        = {ActivityNet: {A} large-scale video benchmark for human activity understanding},
  booktitle    = {Proceedings of the {IEEE} Conference on Computer Vision and Pattern Recognition, {CVPR}
                  2015, Boston, MA, USA, June 7-12, 2015},
  pages        = {961--970},
  year         = {2015}
}

\newpage
\appendix

\section{Implementation Details}

\paragraph{Training Configuration.}
\label{sec:training_setup}
We fine-tune \textbf{LTX-Video~2.3} (22B parameters) from the official base checkpoint on a curated GameVerse subset of \textbf{24{,}380} game-video segments (12--15\,s each) at \textbf{540p / 24\,fps}. Since GameVerse targets visual interactive world modeling, we remove the audio branch of LTX-Video~2.3 prior to fine-tuning and train only the video stream. Training uses a mixed objective with weights \textbf{60\%} image-to-video (i2v), \textbf{20\%} video-to-video (v2v), and \textbf{20\%} text-to-video (t2v), runs for \textbf{3{,}700 steps} at batch size 1 per GPU under FSDP across \textbf{64 GPUs}. Camera conditioning is provided by additive Pl\"ucker injection and action conditioning by \textbf{81-class direction-angle embeddings}. We optimise with AdamW at learning rate $1\times10^{-4}$ on a constant schedule with a 200-step warmup, and retain the base model's RoPE (\texttt{max\_pos}\,$=\,$20) under fractional-position normalisation so that all training samples remain within the position range seen during pretraining.

\paragraph{Data Processing Efficiency.}

To support large-scale data construction, we optimize each stage of the GameVerse pipeline for high-throughput processing. 
For scene-consistent segmentation, we adopt FFmpeg with NVENC acceleration, which significantly improves video decoding and preprocessing efficiency. Combined with TransNetV2, the segmentation pipeline achieves a processing throughput of approximately 400 FPS per worker.
For camera trajectory estimation, we employ a distributed ViPE inference pipeline deployed on 256 GPUs. 
On average, pose estimation for each gameplay video requires approximately 6 minutes, enabling scalable processing of long-duration videos.
For structured semantic annotation, we deploy Kimi-K2.5 using the SGLang inference framework on H20 90GB GPUs. 
The caption generation stage is executed on 16 compute nodes with a total of 128 GPUs, allowing efficient large-scale generation of multi-granular semantic annotations.

\section{Detailed Multi-stage Video Quality Assessment Pipeline}
\label{appendix:quality_pipeline}

Even after segmentation, raw gameplay clips often contain undesirable artifacts such as subtitles, watermarks, UI overlays, compression noise, and unstable motion. These artifacts introduce spurious correlations and degrade temporal consistency, which are particularly harmful for world modeling. To address this, we design a \textbf{multi-stage quality assessment pipeline} with explicit criteria and calibrated thresholds, jointly considering visual quality, content cleanliness, and temporal stability.

\paragraph{Overall Formulation.}
We define the overall quality of a video clip $x$ as a weighted combination of multiple metrics:
\begin{equation}
Q(x) = \sum_k w_k \cdot q_k(x),
\end{equation}
where $q_k(x)$ denotes the $k$-th quality metric and $w_k$ is its corresponding weight. In practice, we adopt a hybrid strategy that combines hard constraints with soft scoring, ensuring both strict filtering and relative ranking.

\paragraph{Visual Quality.}
We first evaluate low-level visual properties to remove clips with obvious degradation. Specifically, we require all clips to satisfy a minimum resolution of 720p, frame rate within $[24, 65]$ fps, and valid encoding formats (e.g., H.264, HEVC, or AV1). We further compute luminance using the BT.601 standard on frames sampled at 5 fps, and discard clips that contain prolonged underexposed or overexposed segments, defined as more than 15 consecutive frames with luminance below 16 or above 240. In addition, we detect border artifacts such as letterboxing and pillarboxing by measuring pixel variance along image boundaries; regions with standard deviation below 4.0 are treated as uniform borders, and clips are removed if such regions occupy more than 10\% of the frame.

\paragraph{Content Cleanliness.}
To eliminate non-world signals, we explicitly filter out text overlays and UI elements. We apply OCR-based detection to identify subtitles and watermarks, focusing on the lower portion of the frame ($y \in [0.7, 1.0]$). Text regions that persist across multiple frames or occupy more than 2\% of the image area trigger rejection, with a minimum detectable text height ratio of 0.008 to capture small HUD elements. In addition, we detect UI overlays based on their temporal stability and spatial structure. Specifically, we compute the temporal standard deviation $\sigma_t$, mean intensity $\mu_t$, and spatial gradient magnitude $|\nabla I|$ for each pixel, and identify UI regions as those satisfying $(\sigma_t < 4) \land (\mu_t > 25) \land (|\nabla I| > 12)$. Clips with UI coverage exceeding 4\% are removed. This step is crucial to prevent models from overfitting to game-specific interface signals.

\paragraph{Temporal Stability.}
We then evaluate whether clips exhibit stable and physically plausible dynamics. Motion is estimated using optical flow (Farneback) at 4 fps, where frames with mean flow magnitude below 0.3 are considered static; clips with more than 50\% such frames are discarded. We further detect freezing artifacts by measuring frame-to-frame luminance differences: consecutive frames with differences below 0.5 are marked as frozen, and clips are rejected if freezing persists for more than approximately 2 seconds or exceeds 30\% of the clip. To capture unstable camera motion, we analyze displacement statistics within sliding windows and identify jitter spikes when instantaneous motion exceeds five times the local mean; clips with more than four such spikes per minute are filtered out.

Beyond these rule-based filters, we incorporate learned quality metrics. We compute multi-dimensional quality scores using COVER, including semantic, technical, and aesthetic components, and retain only the top 70\% of clips ranked by the sum of aesthetic and technical scores. We further evaluate temporal consistency using VBench~\cite{huang2024vbench} metrics, including subject consistency, background consistency, and temporal flickering, with thresholds set to 0.65, 0.85, and 0.83 respectively.

\paragraph{Final Selection.}
The final decision integrates all criteria. Hard constraints (e.g., format validity, OCR detection, UI filtering, motion and freeze rejection) ensure that low-quality samples are strictly removed, while soft scoring mechanisms (e.g., COVER ranking and VBench thresholds) provide finer-grained quality control. Only clips that satisfy all constraints and exceed the required thresholds are retained.

\paragraph{Effectiveness.}
This multi-stage pipeline yields a dataset with high visual clarity, clean semantic content, and stable temporal dynamics. Compared to simple filtering strategies, it provides a more robust and comprehensive quality control mechanism, which is essential for training reliable and generalizable world models.

\section{Additional Experimental Details and Analyses}
\label{appendix:additional}

\subsection{Evaluation Metrics}
\label{appendix:eval_metrics}

\paragraph{VBench.}
We adopt six evaluation dimensions from VBench~\cite{huang2024vbench} to assess temporal consistency, perceptual quality, and semantic alignment of generated videos. \emph{Subject Consistency} measures the stability of subject appearance across frames using DINO feature similarity, capturing identity drift and unintended changes in visual attributes. \emph{Background Consistency} evaluates scene-level coherence through CLIP feature similarity, reflecting whether background structures remain visually consistent over time. \emph{Motion Smoothness} assesses the temporal continuity of generated motion using a video frame interpolation model, penalizing abrupt or unnatural transitions. For frame-level quality, \emph{Aesthetic Quality} measures visual appeal using the LAION aesthetic predictor, considering aspects such as composition and color harmony, while \emph{Imaging Quality} employs MUSIQ to evaluate visual distortions, including blur, noise, and overexposure. Finally, \emph{Overall Consistency} measures semantic and stylistic alignment between generated videos and their textual descriptions using ViCLIP. Together, these metrics characterize complementary aspects of video generation quality, particularly temporal coherence and visual fidelity over extended rollouts.

\paragraph{WorldScore.}
We further employ WorldScore~\cite{duan2025worldscore} to evaluate world generation quality and controllability from geometric, semantic, and perceptual perspectives. The benchmark includes three controllability metrics and four quality metrics. \emph{Camera Control} measures adherence to the target camera trajectory by comparing estimated camera poses from generated videos with the prescribed motion, incorporating scale-invariant rotation and translation errors. \emph{Object Control} evaluates whether prompt-specified objects appear in generated scenes using open-set object detection, while \emph{Content Alignment} measures semantic agreement with the complete input description using CLIPScore. For generation quality, \emph{3D Consistency} evaluates geometric stability across frames through reprojection errors based on DROID-SLAM reconstruction. \emph{Photometric Consistency} assesses the stability of visual appearance and textures using forward--backward optical-flow consistency, capturing artifacts such as texture shifts and flickering. \emph{Style Consistency} evaluates whether the visual style remains consistent by comparing feature Gram matrices across generated frames. Finally, \emph{Subjective Quality} estimates perceptual quality through an automatic combination of CLIP-IQA+ and CLIP Aesthetic scores, rather than direct human ratings.

\subsection{Camera-Pose Supervision Sources}
\label{appendix:pose_sources}

\paragraph{LingBot-World.}
LingBot-World~\cite{robbyant2026lingbot} adopts a hybrid data engine combining general videos, synchronized game recordings, and synthetic Unreal Engine (UE) rendering. Game recordings provide temporally aligned visual observations, user actions, and camera parameters, while UE-rendered sequences offer ground-truth camera intrinsics and extrinsics. For general videos lacking camera annotations, LingBot employs MegaSAM to estimate camera poses. Its camera-conditioning mechanism further incorporates continuous camera information through Pl\"ucker embeddings for fine-grained viewpoint control.

\paragraph{HY-World 1.5.}
HY-World~1.5~\cite{hyworld2025,sun2025worldplay} constructs its training corpus from approximately 320K clips, including 170K AAA-game recordings, 60K reconstructed scenes from DL3DV, 50K synthetic UE-rendered sequences, and 40K real-world videos from Sekai. The UE-rendered subset provides ground-truth camera trajectories, while the DL3DV subset uses 3D reconstruction and customized trajectory rendering for geometrically consistent supervision. For videos without available camera poses, ViPE is used for trajectory estimation and subsequent motion-consistency filtering. HY-World thus combines rendered, reconstruction-derived, and estimated trajectories with dedicated camera-conditioning mechanisms.

\paragraph{Comparison with GameVerse.}
Unlike these systems, GameVerse primarily collects natural gameplay recordings without direct access to game-engine camera parameters. Camera trajectories are recovered using ViPE and subjected to multi-stage quality filtering to remove abrupt jumps, jitter, and unreliable motion. While estimated trajectories lack the precision guarantees of engine-rendered ground truth, they enable scalable coverage of diverse gameplay motions, including irregular navigation, coupled translation and rotation, and complex viewpoint transitions. Therefore, differences in camera-control performance may reflect both pose-supervision sources and conditioning architectures. Nevertheless, improvements from HY-World~1.5 to HY-World-FT demonstrate the utility of GameVerse's estimated trajectories for geometric supervision.

\subsection{Camera-Trajectory Reliability by Motion Type}
\label{appendix:pose_quality}

To assess the reliability of GameVerse's estimated trajectories, we evaluate pose completeness, temporal stability, and cross-estimator agreement across six camera-motion categories: forward, lateral, orbit, tilt, mixed rotation, and complex motion. As shown in Table~\ref{tab:pose_quality}, all audited retained trajectories contain valid poses throughout their duration, with generally low jump rates and rotational jerk. We further compare ViPE trajectories against an independent DROID-SLAM estimator after Sim(3) alignment. Rotation RPE ranges from 0.75 to 1.47 degrees, while translation-direction error ranges from 4.84 to 11.03 degrees, with mixed rotation exhibiting the largest errors. These results support the temporal stability and cross-estimator consistency of the retained trajectories, although they do not establish absolute pose accuracy without ground-truth trajectories. The reported 100\% valid-pose ratios refer to the retained audit set rather than all raw source videos.

\begin{table}[t]
\centering
\small
\setlength{\tabcolsep}{4pt}
\caption{Trajectory quality across six motion types after filtering. Rotation RPE and translation-direction error are computed against DROID-SLAM after Sim(3) alignment. They measure cross-estimator agreement, not ground-truth accuracy. Jump and jerk columns reproduce the source statistics.}
\label{tab:pose_quality}
\begin{adjustbox}{max width=\linewidth}
\begin{tabular}{lccccc}
\toprule
Motion & Valid poses $\uparrow$ & Jump $\downarrow$ & Rot. jerk $\downarrow$ & Rot. RPE ($^\circ$) $\downarrow$ & Trans. dir. ($^\circ$) $\downarrow$ \\
\midrule
Forward & 100.00\% & 0.0000 & 0.0055 & 0.75 & 4.84 \\
Lateral & 100.00\% & 0.0000 & 0.0057 & 1.31 & 6.42 \\
Orbit & 100.00\% & 0.0014 & 0.0099 & 1.31 & 6.59 \\
Tilt & 100.00\% & 0.0006 & 0.0084 & 1.40 & 9.10 \\
Mixed-rot & 100.00\% & 0.0220 & 0.0126 & 1.47 & 11.03 \\
Complex & 100.00\% & 0.0006 & 0.0061 & 0.88 & 4.94 \\
\bottomrule
\end{tabular}
\end{adjustbox}
\vspace{-5pt}
\end{table}

\subsection{Diagnostic Comparison of Original Backbones}
\label{appendix:backbones}

To examine the influence of backbone architectures, we compare the original Wan2.2~\cite{wan2025wan} and LTX-Video~2.3 models without camera conditioning or GameVerse fine-tuning at 5-, 10-, and 15-second generation durations. Both models use the same 103 input images and captions, a resolution of $1024\times576$, seed 42, 40 sampling steps, and a single-stage, full-resolution, non-distilled pipeline. Their native frame rates differ (16\,fps for Wan2.2 and 24\,fps for LTX-Video~2.3), which may affect motion-dependent metrics.

As shown in Table~\ref{tab:backbone}, Wan2.2 consistently achieves higher aesthetic and imaging quality across all three durations, with average advantages of 1.87 and 11.64 points, respectively. These results suggest that appearance-related performance differences between LingBot and LTX-FT may partly originate from their underlying backbones rather than GameVerse fine-tuning alone. However, this diagnostic does not isolate the causes of cross-model differences in 60-second generation. The controlled comparison between HY-World~1.5 and HY-World-FT in the main text provides more direct evidence of the benefits of GameVerse.

\begin{table}[t]
\centering
\small
\setlength{\tabcolsep}{4pt}
\caption{Diagnostic comparison of original backbones without GameVerse fine-tuning or camera conditioning. Both use the same 103 inputs, $1024\times576$ resolution, seed 42 and 40 sampling steps. Wan2.2 uses 16\,fps and LTX-2.3 uses 24\,fps. Scores are multiplied by 100.}
\label{tab:backbone}
\begin{adjustbox}{max width=\linewidth}
\begin{tabular}{llcccccc}
\toprule
Duration & Model & Subj. Cons. & Bkg. Cons. & Mot. Smth. & Aes. Qual. & Img. Qual. & Overall Cons. \\
\midrule
\multirow{2}{*}{5\,s}
& Wan2.2 & 76.72 & 89.91 & 95.87 & 13.30 & 63.97 & 18.89 \\
& LTX-2.3 & 72.86 & 87.36 & 96.35 & 11.13 & 52.36 & 19.31 \\
\midrule
\multirow{2}{*}{10\,s}
& Wan2.2 & 77.69 & 90.15 & 96.92 & 13.45 & 65.30 & 19.34 \\
& LTX-2.3 & 74.18 & 86.97 & 96.78 & 11.89 & 51.25 & 18.71 \\
\midrule
\multirow{2}{*}{15\,s}
& Wan2.2 & 75.13 & 88.86 & 97.46 & 10.04 & 63.54 & 18.75 \\
& LTX-2.3 & 71.37 & 85.90 & 97.52 & 8.16 & 54.29 & 18.81 \\
\bottomrule
\end{tabular}
\end{adjustbox}
\end{table}

\begin{figure}[t]
    \centering
    \includegraphics[width=\linewidth]{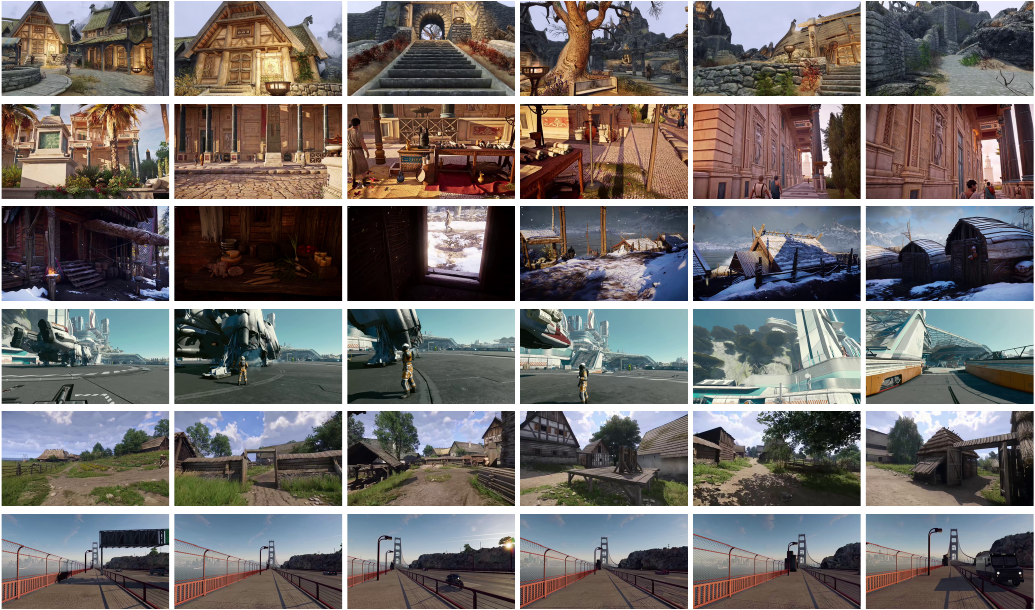}
    \caption{
    Example long-duration gameplay sequences from \textbf{GameVerse}. 
    Each row shows temporally continuous frames sampled from a single gameplay clip across diverse game environments, including indoor exploration, natural scenes, urban driving areas, sunset city environments, and nighttime interaction scenarios. 
    The examples demonstrate the strong temporal continuity, environmental diversity, camera motion consistency, and rich visual dynamics provided by GameVerse.
    }
        \label{fig:video_visual}
\vspace{-3mm}
\end{figure}

\begin{figure*}[th!!]
\centering
\fbox{
\parbox{0.92\textwidth}{
\vspace{1mm}

\textbf{Structured Annotation Format in GameVerse}

\vspace{1mm}

\textbf{[Global Metadata]}
\begin{itemize}
    \item Weather / Time / Season / Lighting
    \item Camera Perspective / Motion / Stability
    \item Environment Type / Architecture Style
    \item Video Style / Visual Quality
    \item Global Scene Description
\end{itemize}

\vspace{1mm}

\textbf{[Segment-level Temporal Annotations]}

Each video is divided into temporally aligned segments, where each segment contains:

\begin{itemize}

\item \textbf{Scene State}
\begin{itemize}
    \item Spatial Layout
    \item Lighting Condition
    \item Ground Surface
    \item Vegetation Density
    \item Crowd Density
    \item Functional Zone
\end{itemize}

\item \textbf{Dynamic Events}
\begin{itemize}
    \item Scene Transition
    \item Motion Change
    \item Lighting Change
    \item Environment Change
    \item Interaction Events
\end{itemize}

\item \textbf{Agent Behaviors}
\begin{itemize}
    \item Ego Action
    \item Ego Description
    \item Pedestrian Flow
    \item NPC Behaviors
    \item Vehicle Presence
\end{itemize}

\item \textbf{Environment Elements}
\begin{itemize}
    \item Entered Elements
    \item Exited Elements
    \item Changed Elements
\end{itemize}

\item \textbf{Scene Description}
\begin{itemize}
    \item Temporally aligned natural-language description of the current segment
\end{itemize}

\end{itemize}

\vspace{1mm}

\textbf{[Annotation Characteristics]}
\begin{itemize}
    \item Multi-granular semantic decomposition
    \item Explicit temporal scene evolution
    \item Camera-aware structured annotations
    \item Interaction-centric dynamic descriptions
    \item Environment-level structural consistency
\end{itemize}

\vspace{1mm}
}
}
\caption{
Overview of the structured semantic annotation format in \textbf{GameVerse}. 
Each gameplay video is decomposed into global metadata and temporally aligned segment-level annotations, jointly modeling scene states, dynamic events, agent behaviors, environmental changes, and temporal scene evolution.
}
\label{fig:annotation_structure}
\vspace{-3mm}
\end{figure*}

\section{Visualization}
\label{appendix:visualization}

Figure~\ref{fig:video_visual} presents representative long-duration gameplay sequences from GameVerse.
Each row corresponds to temporally continuous frames sampled from a single gameplay clip, illustrating the diversity of environments, lighting conditions, gameplay perspectives, and interaction dynamics contained in the dataset.
The examples cover indoor exploration scenes, open-world natural environments, urban roads, sunset cityscapes, and nighttime gameplay scenarios across multiple game titles.
Notably, the sampled sequences exhibit strong temporal continuity and stable camera motion over long horizons, while preserving diverse scene layouts and environmental transitions. These examples further demonstrate the ability of GameVerse to capture complex world dynamics and long-term scene evolution for interactive world modeling.

\section{Four-level Structured Semantic Annotations}
\label{appendix:caption_details}

To better support long-horizon interactive world modeling, GameVerse introduces a four-level structured semantic annotation framework that explicitly decomposes gameplay videos into complementary semantic components.
Unlike conventional video captions that describe an entire clip using a single sentence, our annotations separately model \textbf{global static}, \textbf{local static}, \textbf{dynamic events}, and \textbf{agent actions}, enabling more structured supervision for controllable generation and temporal world understanding.

Figure~\ref{fig:annotation_structure} summarizes the overall annotation format of GameVerse.
Each gameplay video is first associated with global metadata describing environmental conditions, camera properties, and high-level scene semantics.
The video is then divided into temporally aligned segments, where each segment contains structured annotations for local scene structure, dynamic events, interaction behaviors, environmental evolution, and natural-language scene descriptions.

\begin{table*}[t]
\centering
\caption{
Example annotation from a historical urban environment (Assassin's Creed-style gameplay).
}
\label{tab:appendix_case1}
\renewcommand{\arraystretch}{1.15}
\setlength{\tabcolsep}{5pt}
\resizebox{\textwidth}{!}{
\begin{tabular}{p{3cm} p{13cm}}
\toprule

\textbf{Category} & \textbf{Annotation Example} \\

\midrule

Global Description
&
Third-person gameplay footage showing a hooded protagonist exploring a historical Middle Eastern city with arcades, markets, plazas, fountains, and crowded urban streets.
\\

Dynamic Environment 
&
Clear weather, midday lighting, historic sandstone architecture, pedestrian streets, moderate crowd density.
\\

Scene Evolution
&
Covered stone arcade $\rightarrow$ narrow residential street $\rightarrow$ market district $\rightarrow$ central plaza $\rightarrow$ elevated garden terrace.
\\

Dynamic Events
&
Exiting covered arcade, navigating market crowds, interacting with NPC flow, turning around central fountain, ascending garden stairs.
\\

Agent Behaviors
&
Walking through crowded streets, exploring urban plazas, navigating around market vendors and animals.
\\

Key Scene Elements
&
Stone archways, fabric awnings, domed buildings, market stalls, fountains, palm trees, flower beds, garden terraces.
\\

\bottomrule
\end{tabular}
}
\vspace{-2mm}
\end{table*}

\begin{table*}[t]
\centering
\caption{
Example annotation from a \textbf{third-person cinematic exploration} sequence.
}
\label{tab:annotation_example}
\renewcommand{\arraystretch}{1.15}
\setlength{\tabcolsep}{5pt}
\resizebox{\textwidth}{!}{
\begin{tabular}{p{3.2cm} p{12.8cm}}
\toprule

\textbf{Category} & \textbf{Annotation Example} \\

\midrule

Global Description
&
Third-person gameplay footage showing a warrior character exploring a rocky desert environment, passing through cave tunnels, ancient ruins, and lakeside areas under dynamic lighting conditions.
\\

Weather / Time / Style
&
Clear weather, midday lighting, cinematic visual style, third-person follow camera.
\\

Camera Motion
&
Walking-forward camera motion with stabilized chest-level perspective.
\\

Scene Segment 1
&
Wide desert landscape with large rock formations and sparse vegetation. 
The character walks toward a distant cave entrance while lightning flashes in the background.
\\

Scene Segment 2
&
The character enters a narrow cave tunnel, transitioning from bright sunlight into a dark enclosed environment with rough stone walls.
\\

Scene Segment 3
&
The scene opens into ancient ruins dominated by a massive tree structure, surrounded by stone pillars and distant mountains.
\\

Dynamic Events
&
Lightning strikes, scene transitions, entering enclosed spaces, approaching environmental landmarks.
\\

Agent Actions
&
Walking forward, exploring ruins, traversing shoreline areas, entering shallow water.
\\

Environment Elements
&
Ancient ruins, cave archways, mountain ranges, lakeside shoreline, rocky terrain, stone pillars.
\\

\bottomrule
\end{tabular}
}
\vspace{-2mm}
\end{table*}
\begin{table*}[t]
\centering
\caption{
Example annotation from a \textbf{first-person cinematic exploration} sequence.
}
\label{tab:appendix_case3}
\renewcommand{\arraystretch}{1.15}
\setlength{\tabcolsep}{5pt}
\resizebox{\textwidth}{!}{
\begin{tabular}{p{3cm} p{13cm}}
\toprule

\textbf{Category} & \textbf{Annotation Example} \\

\midrule

Global Description
&
First-person gameplay footage showing exploration of coastal ruins, medieval stone architecture, beaches, and underground cave systems.
\\

Dynamic Environment 
&
Partly cloudy weather, cinematic rendering, first-person camera, coastal environment, mixed indoor-outdoor transitions.
\\

Scene Evolution
&
Rocky canyon $\rightarrow$ covered corridor $\rightarrow$ open beach $\rightarrow$ castle entrance $\rightarrow$ underground cave.
\\

Dynamic Events
&
Transition from canyon to open waterfront, entering covered structures, discovering cave interiors illuminated by natural light shafts.
\\

Agent Behaviors
&
Walking through environmental transitions, approaching landmarks, exploring caves and hidden interiors.
\\

Key Scene Elements
&
Stone bridges, water canals, medieval castle, armored statues, cave grotto, large underground tree.
\\

\bottomrule
\end{tabular}
}
\vspace{-2mm}
\end{table*}
\begin{table*}[t]
\centering
\caption{
Example annotation from a \textbf{large-scale natural exploration} environment.
}
\label{tab:appendix_case2}
\renewcommand{\arraystretch}{1.15}
\setlength{\tabcolsep}{5pt}
\resizebox{\textwidth}{!}{
\begin{tabular}{p{3cm} p{13cm}}
\toprule

\textbf{Category} & \textbf{Annotation Example} \\

\midrule

Global Description
&
Third-person gameplay footage of a warrior character exploring rocky deserts, ancient ruins, caves, and lakeside environments under cinematic lighting.
\\

Dynamic Environment 
&
Clear weather, desert terrain, natural sunlight, wide-open spatial layout, sparse vegetation.
\\

Scene Evolution
&
Rocky desert $\rightarrow$ cave tunnel $\rightarrow$ ancient ruins $\rightarrow$ lakeside shoreline.
\\

Dynamic Events
&
Lightning strikes in the distance, entering enclosed cave areas, approaching monumental ruins, transitioning toward waterfront scenery.
\\

Agent Behaviors
&
Exploring ruins, traversing cave systems, approaching environmental landmarks, entering shallow water.
\\

Key Scene Elements
&
Rock formations, cave entrances, ancient ruins, giant world tree, mountain ranges, lakeshore environments.
\\

\bottomrule
\end{tabular}
}
\vspace{-2mm}
\end{table*}

\paragraph{Global Static.}
The first level focuses on globally consistent environmental properties throughout the video sequence.
These annotations describe high-level scene attributes such as environment type, architectural style, weather conditions, lighting conditions, season, camera perspective, and overall scene composition.
As shown in Tables~\ref{tab:appendix_case1}--\ref{tab:appendix_case3}, the annotations cover highly diverse environments including historical Middle Eastern cities, rocky desert landscapes, coastal ruins, underground caves, and large-scale natural environments.
Such annotations provide stable world-level context and coarse-grained environmental priors for long-horizon generation.

\paragraph{Local Static.}
The second level captures fine-grained spatial structures within temporally aligned segments.
Compared with global context annotations, this level focuses on local scene layouts and environment details that evolve over time.
For example, the annotations explicitly describe transitions between covered stone arcades, narrow residential streets, open plazas, cave tunnels, lakeside shorelines, and underground grotto environments.
These local structural descriptions preserve spatial continuity and geometric organization across long-duration gameplay sequences.

\paragraph{Dynamic Events.}
The third level models temporally evolving events and environmental transitions.
Instead of only describing static scenes, the annotations explicitly capture scene evolution over time, including entering enclosed caves, exiting urban districts, approaching environmental landmarks, navigating through crowded markets, and transitioning between indoor and outdoor environments.
As illustrated in the examples, GameVerse also captures environmental events such as lightning flashes, motion changes, and camera-view transitions, enabling richer supervision for temporal causality and long-horizon scene evolution.

\paragraph{Agent Actions.}
The fourth level focuses on interaction-centric behaviors of the player character and surrounding agents.
These annotations describe actions such as walking, exploration, climbing, stealth movement, driving, environmental interaction, and NPC activities.
For example, the examples include navigating crowded urban streets, traversing cave systems, approaching historical landmarks, and exploring large-scale outdoor environments.
Compared with conventional action labels, the proposed annotations jointly model both the action itself and its surrounding interaction context, enabling richer supervision for controllable interaction modeling and embodied world simulation.

Together, these four semantic levels provide complementary supervision signals for long-horizon video understanding and generation. By explicitly separating global context, local structure, dynamic events, and interaction behaviors, GameVerse provides substantially richer annotations than conventional flat captioning approaches and better supports controllable interactive world modeling.

\begin{figure}[h]
    \centering
    \includegraphics[width=\linewidth]{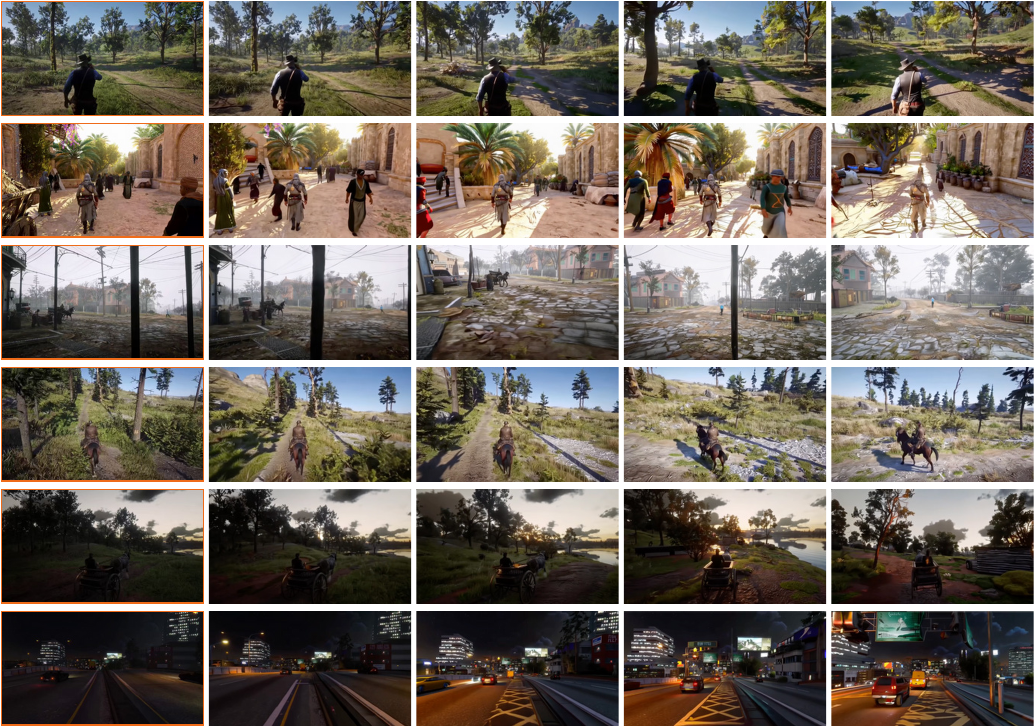}
    \caption{
    Qualitative image-to-video generation results of the LTX-2.3 model fine-tuned on \textbf{GameVerse}. 
    Given a single initial frame together with camera trajectories and structured semantic captions, the model generates long-duration gameplay videos with consistent scene layouts, stable camera motion, and coherent temporal dynamics across diverse environments, including natural exploration, urban interaction, horse riding, and nighttime driving scenarios.
    }
    \label{fig:case_study_vbench}
\vspace{-3mm}
\end{figure}

\section{Case Study}
\label{appendix:case_study}

\subsection{Qualitative Long-horizon Generation.}

Figure~\ref{fig:case_study_vbench} presents representative image-to-video generation results using the LTX-Video-2 model fine-tuned on GameVerse. 
Given only a single initial frame, together with the corresponding camera trajectories and structured semantic captions, the model is able to generate temporally coherent long-duration gameplay videos across diverse environments and interaction scenarios.

The generated sequences demonstrate strong scene consistency and stable camera-aware motion over long temporal horizons. 
For example, the model successfully preserves environmental geometry and spatial layouts in large-scale outdoor exploration scenes, while maintaining smooth camera transitions and consistent illumination changes in urban streets, horseback riding, and nighttime driving scenarios. 
Moreover, the generated videos exhibit coherent interaction dynamics and realistic environmental evolution, indicating that the structured semantic supervision provided by GameVerse effectively improves long-horizon controllable video generation.

These qualitative results further demonstrate the effectiveness of GameVerse for geometry-aware, temporally consistent, and interaction-centric world modeling.

\begin{figure}[h]
    \centering
    \includegraphics[width=\linewidth]{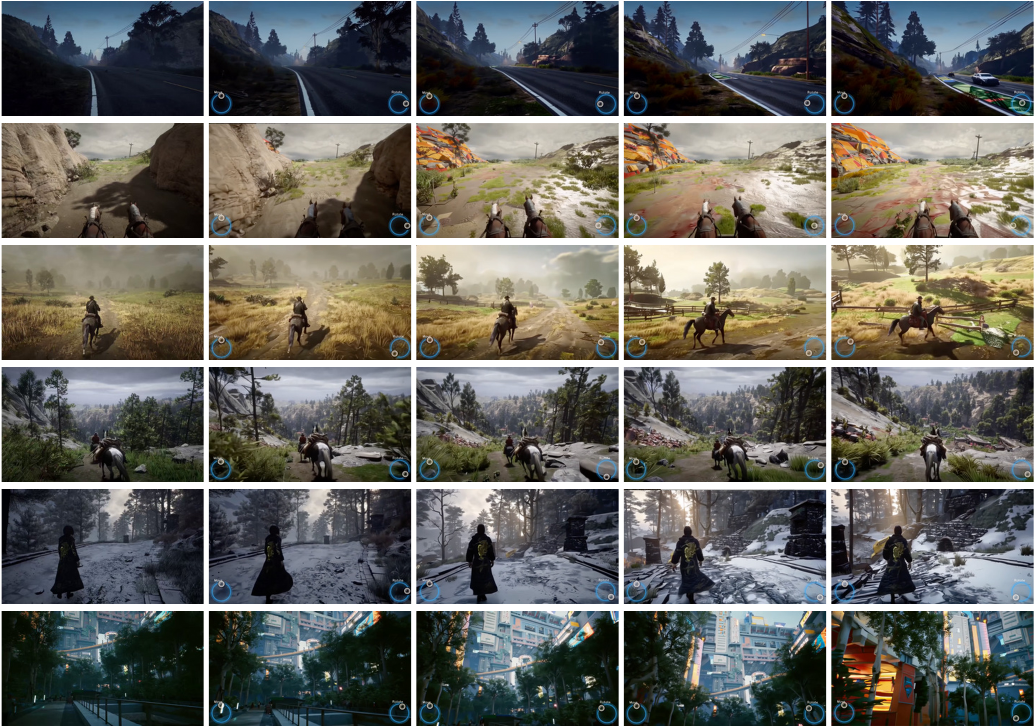}
\caption{
Qualitative image-to-video generation results conditioned on camera trajectories and structured semantic captions. 
The blue trajectory overlays visualize the input camera motion signals used for controllable generation. 
Given a single initial frame, the LTX-Video-2 model fine-tuned on \textbf{GameVerse} is able to generate temporally coherent long-duration gameplay videos with stable camera-aware motion, consistent scene geometry, and smooth environmental transitions across diverse gameplay scenarios.
}
        \label{fig:case_study_pose}
\vspace{-3mm}
\end{figure}

\subsection{Qualitative Camera-Controlled Generation.}

Figure~\ref{fig:case_study_pose} presents qualitative image-to-video generation results conditioned on both camera trajectories and structured semantic captions. 
The overlaid blue trajectories visualize the input pose control signals provided to the generation model.

As shown in the examples, the model is able to follow complex camera motions while preserving long-term scene consistency and coherent environmental structure. 
Across diverse gameplay scenarios, including road navigation, horseback riding, forest exploration, snowy environments, and urban scenes, the generated videos exhibit smooth camera-aware motion and stable temporal evolution over long horizons.

Moreover, the generated sequences maintain consistent scene geometry and spatial layouts under large camera movements, suggesting that the camera trajectory supervision provided by GameVerse effectively improves geometry-aware and controllable long-duration video generation.

\newpage

\end{document}